\documentclass{article}

\usepackage{microtype}
\usepackage{graphicx}
\usepackage{subfigure}
\usepackage{booktabs} % for professional tables

\usepackage{hyperref}

\usepackage[accepted]{icml2025}

\usepackage{amsmath}
\usepackage{amssymb}
\usepackage{mathtools}
\usepackage{amsthm}

\usepackage[capitalize,noabbrev]{cleveref}

\usepackage{xspace}
\usepackage{wrapfig}
\usepackage{dirtytalk}
\usepackage{tcolorbox}
\usepackage{enumitem}
\newcommand{\alg}{P3O\xspace}
\newcommand{\prpo}{PRPO\xspace}

\newcommand{\dpo}{DPO\xspace}

\newcommand{\alglong}{Pessimistic Preference-based Policy Optimization\xspace}
\newcommand{\prpolong}{Pessimistic Reward-based Policy Optimization\xspace}
\tcbset{
    graybox/.style={
        colback=gray!20,  % Background color (light gray)
        colframe=black,   % Border color (black)
        boxrule=1pt,      % Border thickness
        arc=4mm,          % Box corner rounding
        width=\textwidth, % Box width to match the text width
        before=\vspace{10pt}, % Space before the box
        after=\vspace{10pt},  % Space after the box
        center             % Center the box
    }
}
\def\Re{\mathbb{R}}

\def\Nat{{\rm I\kern\pIR N}}
\def\argmax{\mathop{\rm arg\,max}}
\def\argmin{\mathop{\rm arg\,min}}

\def\log{\mathop{\rm log}}

\newcommand{\EE}[1]{\exptE\left[#1\right]}

\def\D{{\mathcal{D}}}
\def\E{{\mathcal{E}}}

\def\L{{\mathcal{L}}}

\def\P{{\mathcal{P}}}

\def\R{{\mathcal{R}}}

\def\X{{\mathcal{X}}}
\def\Y{{\mathcal{Y}}}

\def\vec0{{\boldsymbol{0}}}

\def\vecy{{\boldsymbol{y}}}

\newcommand{\beq}{\begin{equation}}
\newcommand{\eeq}{\end{equation}}
\newcommand{\beqa}{\begin{eqnarray}}
\newcommand{\eeqa}{\end{eqnarray}}
\newcommand{\beqan}{\begin{eqnarray*}}
\newcommand{\eeqan}{\end{eqnarray*}}
\newcommand{\ben}{\begin{eqnarray*}}
\newcommand{\een}{\end{eqnarray*}}

\renewcommand{\EE}[2]{\mathbb{E}_{#1\!\!}\left[#2\right]}

\def\E#1{\EE{\,}{#1}}

\usepackage{cancel}
\usepackage[normalem]{ulem}

\renewcommand{\epsilon}{\varepsilon}

\renewcommand{\E}{\ensuremath{\mathbb{E}}}

\usepackage{xcolor}

\newcommand{\piref}{\ensuremath{\pi_{\mathrm{ref}}}}
\newcommand{\KL}[2]{{\mathrm{KL}}\left(#1\|#2\right) }

\newcommand{\pimix}{\pi_{\mathrm{mix}}}
\newcommand{\pimixalpha}{\pimix^\alpha}
\newcommand{\pisample}{\pi_{\mathrm{data}}}
\newcommand{\pinash}{\pi_{\mathrm{nash}}}
\newcommand{\pbt}{p_{\mathrm{BTL}}}
\newcommand{\pirlhf}{\pi_{\mathrm{rlhf}}}
\newcommand{\pipessrlhf}{\pi_{\mathrm{p}\textrm{{-}}\mathrm{rlhf}}}
\newcommand{\pipessnash}{\pi_{\mathrm{p}\textrm{{-}}\mathrm{nash}}}
\newcommand{\pirespessnash}{\pi_{\mathrm{rp}\textrm{{-}}\mathrm{nash}}}
\newcommand{\piresnash}{\pi_{\mathrm{r}\textrm{{-}}\mathrm{nash}}}

\newcommand{\pmle}{p_{\mathrm{mle}}}

\newcommand{\rmle}{r_{\mathrm{mle}}}
\newcommand{\Lpref}{\L_{\mathrm{pref}}}

\newcommand{\Jrlhf}{J_{\mathrm{R}}}
\newcommand{\Jpref}{J_{\mathrm{P}}}   
\newcommand{\Jpppo}[1]{J_{\mathrm{P3O}(#1)}}   
\newcommand{\Jprpo}[1]{J_{\mathrm{PRPO}(#1)}}  
\newcommand{\cP}{\P} 

\newcommand{\cW}{\ensuremath{\mathcal{W}}}

\renewcommand{\paragraph}[1]{\textbf{#1}~}
\newcommand{\iidsim}{\stackrel{\mathrm{i.i.d.}}{\sim}}

\theoremstyle{plain}
\newtheorem{theorem}{Theorem}[section]

\newtheorem{lemma}[theorem]{Lemma}
\newtheorem{definition}[theorem]{Definition}

\theoremstyle{remark}

\usepackage[textsize=tiny]{todonotes}

\icmltitlerunning{Mitigating Preference Hacking in Policy Optimization with Pessimism} % for \\Robust Alignment from  Preferences}

\begin{document}

\twocolumn[
\icmltitle{Mitigating Preference Hacking in Policy Optimization with Pessimism}% for \\Robust Alignment from  Preferences}


\icmlsetsymbol{equal}{*}

\begin{icmlauthorlist}
\icmlauthor{Dhawal Gupta}{a}
\icmlauthor{Adam Fisch}{b}
\icmlauthor{Christoph Dann}{c}
\icmlauthor{Alekh Agarwal}{c}
\end{icmlauthorlist}

\icmlaffiliation{a}{UMass Amherst}
\icmlaffiliation{b}{Google DeepMind}
\icmlaffiliation{c}{Google Research}

\icmlcorrespondingauthor{Dhawal Gupta}{dhawgupta@gmail.com}
\icmlcorrespondingauthor{Adam Fisch}{fisch@google.com}
\icmlcorrespondingauthor{Christoph Dann}{chrisdann@google.com}
\icmlcorrespondingauthor{Alekh Agarwal}{alekhagarwal@google.com}

\icmlkeywords{Machine Learning}

\vskip 0.3in
]



\printAffiliationsAndNotice{}  % leave blank if no need to mention equal contribution

\begin{abstract}
This work tackles the problem of overoptimization in reinforcement learning from human feedback (RLHF), a prevalent technique for aligning models with human preferences. RLHF relies on reward or preference models trained on \emph{fixed preference datasets}, and these models are unreliable when evaluated outside the support of this preference data, leading to the common reward or preference hacking phenomenon. We propose novel, pessimistic objectives for RLHF which are provably robust to overoptimization through the use of pessimism in the face of uncertainty, and design practical algorithms, P3O and PRPO, to optimize these objectives. Our approach is derived for the general preference optimization setting, but can be used with reward models as well. We evaluate P3O and PRPO on the tasks of fine-tuning language models for document summarization and creating helpful assistants, demonstrating remarkable resilience to overoptimization.
\end{abstract}

\section{Introduction}
Reinforcement learning (RL) from human feedback (RLHF)~\citep{christiano2017deep} has emerged as a promising technique for aligning language models with human preferences~\citep{stiennon2020learning, ouyang2022training}. The predominant approach involves training a reward model on human preference data and then fine-tuning the language model to maximize the expected reward of the responses it generates to training inputs. More recently, a line of work~\citep{swamy2024minimaximalist, munos2023nash, calandriello2024human, guo2024direct} has argued for the benefits of learning a \emph{pairwise} preference function from the preference dataset, and using this to directly compare  trajectories side-by-side during online RL. Irrespective of whether we use reward or preference models during training, however, the availability of a limited pool of high-quality preference data presents a key bottleneck in learning good policies. The high cost of collecting preference datasets with human feedback means that they suffer from limited coverage, and the reward or preference models trained on such datasets fail to adequately generalize to policies which produce trajectories out of the support of the preference data.

The inadequacy of learned reward/preference models in reliably producing good policies has resulted in the now well-documented phenomenon of reward hacking or overoptimization~\citep{amodei2016concrete, gao2023scaling, eisenstein2024helping}. Perhaps the most commonly used technique to limit overoptimization is through regularization by the KL-divergence between the policy being trained and a reference model. However, KL-divergence only measures the distributional distance of the produced responses, independently of the uncertainty in the predictions of the reward/preference models. This typically leads to  either overoptimization or overly limiting reward/preference maximization, when too little or too much regularization is applied, necessitating careful tuning. Even when the trade-off is properly tuned, KL regularization is still often observed to inhibit learning.
\begin{figure*}[th!]
    % \centering
    % \begin{minipage}{0.5\textwidth}
    \centering
    \hspace{0.5cm}
    \includegraphics[height=1.8in]{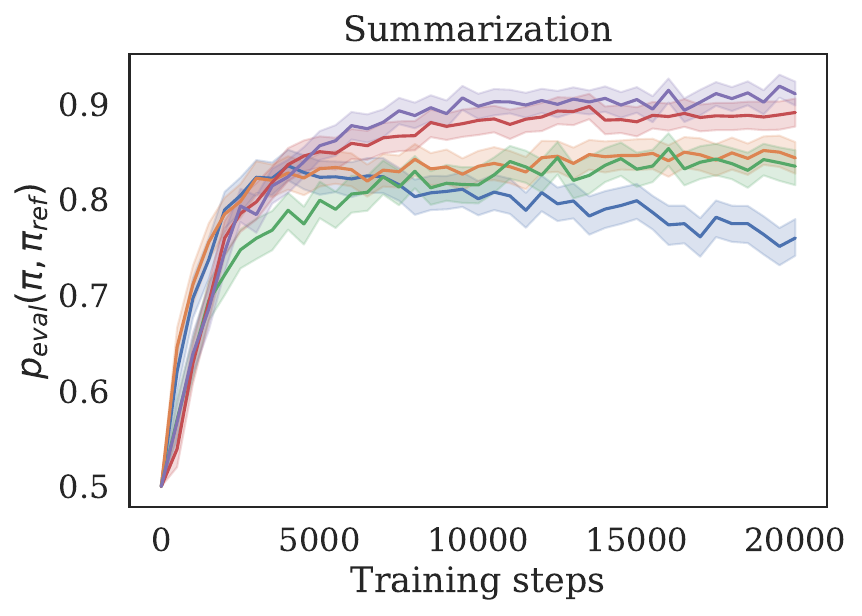}
    \hspace{0.5cm}
    \includegraphics[height=1.8in]{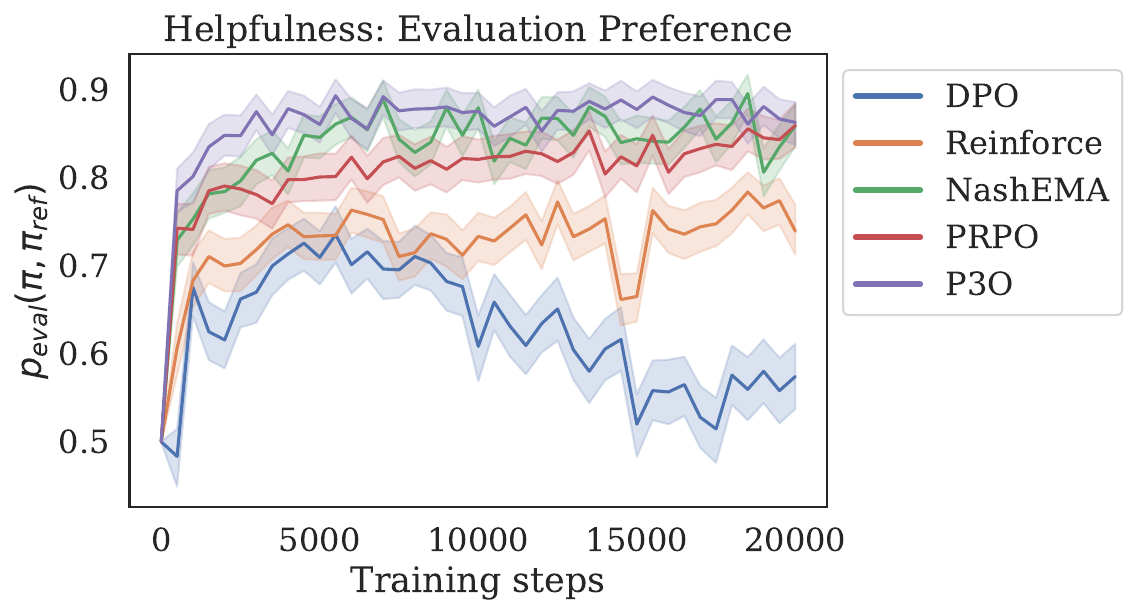}
    % \end{minipage}%
    % \begin{minipage}{0.4\textwidth}
    \vspace{-5pt}
    \caption{ Comparison of our methods (P3O and PRPO) against standard approaches (REINFORCE and DPO) on summarization and ``helpful assistant'' tasks, showing the evaluation preference of a prompted Gemini evaluator, for generations of the policy over those of the reference policy. The hyperparameters of each method have been tuned to prevent reward hacking (best eval performance), which necessitates strong KL-regularization for DPO and REINFORCE. %We observe that this limits them relative to our approaches, which 
    Our approaches, however, can avoid reward hacking by relying on pessimism instead of  KL regularization (a blunt tool), and achieve consistently better performance as a result. We also compare to Nash-EMA~\cite{munos2023nash}, a natural  baseline that also employs a preference model instead of a reward model, but without any pessimism.   Like REINFORCE, the best Nash-EMA values still plateau to a lower performance on summarization. While it achieves similar win-rates over $\piref$ on helpfulness, it does so at the cost of much longer and idiosyncratic generations on average (see Figure~\ref{fig:qualitative-results}). Shaded areas show $95\%$ CIs of the evaluation.\looseness=-1
    }
    \label{fig:eval_pref_gemini}
    \vspace{-5pt}
    % \end{minipage}
\end{figure*}

Motivated by these concerns, there is a growing literature on techniques to control this overoptimization behavior in more data-driven ways, such as by incorporating uncertainty in the predictions of the underlying reward model with explicit reward ensembles~\citep{eisenstein2024helping, coste2023reward}, or pessimistic reasoning~\citep{fisch2024robust, liu2024provably, huang2024correctinga, cen2024valueincentivized}, albeit with  only modest success in standard settings. Similar to the related area of distributionally robust optimization~\cite{Bertsimas:PriceRob, bental2013}, a core part of the challenge is balancing sufficient reward uncertainty and pessimism to prevent overoptimization, while still being able to learn effectively \cite{eisenstein2024helping, fisch2024robust}.

In contrast to the reward-based setting,  much less work has studied the incorporation of uncertainty when using learned pairwise preference models in subsequent RL---with even fewer works devoted to understanding exactly what \emph{kinds} of pessimistic possibilities due to uncertainty are useful to entertain, versus ignore. Pessimistic techniques for preference-based RL are particularly challenging, since many leading methods~\citep{munos2023nash, swamy2024minimaximalist, calandriello2024human} without pessimistic reasoning, critically leverage the symmetry of the min and max players in preference-based RL to develop efficient algorithms, but the introduction of additional pessimism breaks this symmetry. Additionally, the obvious pessimistic estimators studied theoretically in~\cite{cui2022when, ye2024online} exhibit some pathologies when the offline dataset has systematic gaps in its coverage, arising from the non-transitivity of general preferences. These pathologies mean that despite using pessimism, we cannot easily compete with policies whose responses are adequately covered in the data---the typical benchmark for offline RL in reward-based settings.\looseness=-1

%, in large part due to inherent trade-offs between over- and \emph{under}-optimization when the uncertainty is too large to be helpful. 

In this work, we build on prior works in preference-based RLHF~\citep{swamy2024minimaximalist, munos2023nash}, as well as offline learning in Markov games~\citep{cui2022when}, to obtain new robust objectives for incorporating a \emph{restricted} form of uncertainty from finite preference datasets. Specifically, we  make the following key contributions:
\begin{enumerate}[leftmargin=*]%, itemsep=0pt, topsep=0pt]
    \item We identify problematic properties of prior pessimistic estimators~\citep{cui2022when, ye2024online} in the absence of prohibitive coverage assumptions on the preference data sampling policy. We then develop a new \emph{restricted Nash formulation} under which the learned policy is provably preferable to any other competitor policy that is restricted to choosing actions within the support of the preference data sampling policy, and show the theoretical benefits of this formulation.\looseness=-1 
    
    \item We provide a practical algorithm, \alglong~(\alg) for optimizing the resulting objective. We approximate the ideal theoretical objective with a variational upper bound, that yields a minimax game between a policy and a preference player, which we solve using gradient ascent-descent. The policy optimization is similar to prior works~\citep{swamy2024minimaximalist, munos2023nash} and the preference updates are adversarial to the current policy's choices. For Bradley-Terry-Luce models with reward-based preferences, we also evaluate a simpler, reward-based variant  of \alg (which we call \prpo) in our experiments.
    
    \item Empirical evaluation on document summarization and training helpful assistants in Figure~\ref{fig:eval_pref_gemini} shows \alg and \prpo reach a higher quality of responses quickly, and the quality does not degrade due to overoptimization from further training. This is contrast with standard RLHF methods (\dpo~\citep{rafailov2023direct}, and REINFORCE~\citep{willams1992simple, ahmadian2024back}, which either exhibit significant overoptimization, or are limited in their ability to sufficiently optimize. The evaluations are performed using a prompted Gemini 1.5 Flash auto-evaluator~\citep{geminiteam2024geminifamilyhighlycapable}. Detailed analysis in Section~\ref{sec:experiments} further shows that P3O and PRPO avoid qualitative reward hacking behavior of REINFORCE and \dpo. 
\end{enumerate}

\newcommand{\vecyw}{\vecy_{\text{W}}}
\newcommand{\vecyl}{\vecy_{\text{L}}}

\section{Background}

% Notation
We consider human alignment of a language model (LM) policy $\pi\in\Pi$, where $\Pi \subseteq \{ \X \rightarrow \Delta(\Y)\}$\footnote{We use $\Delta(\Y)$ to denote the probability simplex defined over the elements of the set $\Y$ (e.g., the set of possible LM responses).}, which generates for a context $x \in \X$, 
% sampled from $x\sim \rho \in \Delta(\X)$, 
a response $y \sim \pi(\cdot | x)$ with $y \in \Y$.
% Here, $\V$ is the vocabulary of tokens, and $L,H \in \natural$ are the lengths of the context and response, respectively.
We assume that we are given access to a preference dataset, $\D$, consisting of tuples $(x, y_w, y_l) \in \X\times\Y\times\Y$ where for context $x$, the response $y_w$ is preferred over $y_l$ (as labeled by a human). We further assume access to a reference policy $\piref$, which may or may not match the original sampling policy for the preferred and dispreferred $(y_w, y_l)$ responses in $\D$. For brevity, we drop the context $x$ from the notation and work with a finite $\Y$ when there is no ambiguity.\looseness=-1

%\newcommand{\vecthetasft}{\vectheta_{\text{SFT}}}
%\subsection{Preliminaries}
Preferences are often modeled via a reward function under the Bradley-Terry-Luce (BTL)~\citep{bradley1952rank, luce2012individual} model~\citep{christiano2017deep, ouyang2022training}; however, in this paper, we make no such assumptions and work with both general pairwise preference functions as well as BTL preference functions based on pointwise rewards. In the following, we first set up the preference learning framework, and then discuss techniques to optimize policies with preference feedback, while also establishing the use of pessimism to handle uncertainties that may exist in the reward and preference functions.

\paragraph{Learning preferences from data.} We define the preference function $p:\Y\times\Y \rightarrow [0,1]$, such that $p( y_1, y_2) \doteq \Pr(y_1 \succ y_2)$ represents the probability of the generation $y_1$ being preferred over $y_2$ by a target user. Being a probability, the preference function satisfies: $p( y_1, y_2) = 1 - p(y_2, y_1)$.
% We learn a preference model by parameterizing the preference function through parameters $\vecphi \in \Phi$, and represent the parameterized model as $p_\vecphi$. 
To obtain a preference model, we typically fine-tune a pretrained language model (LM) on $\D$ to produce the maximum likelihood estimate $\pmle$ via the following objective:
\begin{align}\label{eq:pref-mle}
\begin{split}
\textstyle    & \pmle \in \argmin_{p} \Lpref(p; \D), ~~\mbox{where}~~
\\ &\Lpref(p; \D) = -\E_{(y_w,y_l) \sim \D}\left[   \log p \left(  y_w, y_l \right)\right].
\end{split}
\end{align}

We overload the notation to say $p(\pi, \pi')$,
%and $r(\pi)$, 
where $\pi, \pi' \in \Pi$, to represent the expected preference for $\pi$ over $\pi'$, given the preference function $p$, that is,
$$
    p(\pi, \pi') = \E_{y \sim \pi, y' \sim \pi'}[p(y, y')].
$$

In the standard case of RLHF, the preference function is modeled using a reward function $r: \Y \rightarrow \Re$, assuming an underlying  BTL model $\pbt$, i.e., 
\begin{align}\label{eq:bt-preference}
    \pbt(y_1, y_2;r) \doteq \frac{\exp(r(y_1))}{\exp(r(y_1)) + \exp(r(y_2))}.
\end{align}
% Then minimizing \eqref{eq:pref-mle} for $\pbt$ gives rise to the reward function $\rmle$. 
Note that the set of BTL models is a strict subset of the general pairwise preference models. We also overload the notation $\pbt(r)$ to denote the BTL preference model that is induced by reward function $r$.

\paragraph{Preference-based policy optimization.} To optimize general preferences without making a BTL modeling assumption, following \citet{munos2023nash} and \citet{swamy2024minimaximalist}, we formulate a preference game $\Jpref(\pi, \pi', p)$ between a pair of competing policies $\pi$ and $\pi'$, with preference function $p$, a reference policy $\piref$, and a regularization parameter $\beta > 0$, as\looseness=-1
\begin{align*}
    \Jpref(\pi, \pi', p) \doteq p(\pi, \pi') - \beta \KL{\pi}{\piref} + \beta \KL{\pi'}{\piref},
\end{align*}
where $\KL{\pi}{\piref} \doteq \E_{y \sim \pi}\left[\log\tfrac{\pi(y)}{\piref(y)}\right]$.
For the preference objective $\Jpref$, the $\pi$ and $\pi'$ players optimize their corresponding $\max$-$\min$ and $\min$-$\max$ objectives, i.e., 
\begin{align}
\begin{split}
    &\pinash \in \argmax_{\pi \in \Pi} \min_{\pi' \in \Pi} \Jpref(\pi, \pi', \pmle),\\ 
    &\pinash' \in \argmin_{\pi' \in \Pi} \max_{\pi \in \Pi} \Jpref(\pi, \pi', \pmle).
    \label{eq:nash}
\end{split}
\end{align}
% where $\Pi \subseteq \{\X \rightarrow \Delta(\Y)\}$ is the class of policies that we optimize over.
Here, due to the symmetry of the game, a Nash equilibrium exists at the same policy, i.e., $\pinash = \pinash'$, and the objective can be simplified to optimizing a \emph{single}-player game---termed Self-play Preference Optimization (SPO) in \citet{swamy2024minimaximalist}.\looseness=-1

Alternatively, in the standard \emph{reward}-based setup, given a reward function $r$,
% after learning the reward function $\rmle$, 
the corresponding objective for reward optimization becomes $\Jrlhf(\pi, r)$, which is defined as:
\begin{align}\label{eq:obj-rlhf}
    \Jrlhf(\pi, r) \doteq \E_{y\sim\pi}[r(y)] - \beta \KL{\pi}{\piref}.
\end{align}
As described earlier, typical reward-based RLHF settings make use of a learned reward $\rmle$ obtained by optimizing \eqref{eq:pref-mle} for $\pbt$ in \eqref{eq:bt-preference}, and fine-tune the policy to maximize the expected pointwise reward, i.e.,\footnote{Note that, as shown by \citet{azar2023general}, this is also equivalent to optimizing $\Jpref$ with a fixed opponent $\pi'$ and a special transformation of the BTL preference function, namely, $p(y_1, y_2) = \Psi(\pbt(y_1, y_2))$ where $\Psi(q) = \log(q / (1 - q))$.}
\begin{align} \label{eq:pi-rlhf}
    &\pirlhf \in \argmax_{\pi \in \Pi} \Jrlhf(\pi, \rmle),\\ &~~\mbox{where}~~
    \rmle \in \argmin_r \Lpref\Big(\pbt(r); \D\Big). \nonumber
\end{align}

\paragraph{Pessimism in preference-based policy optimization.} 
It is a well-understood issue in preference optimization and RLHF that optimizing $\Jpref(\cdot, \cdot, \pmle)$ and $\Jrlhf(\cdot, \rmle)$ can lead to over-optimization of the corresponding preference and reward functions~\citep{gao2023scaling, eisenstein2024helping}. This behavior arises because $\pmle$ (resp. $\rmle$) has large inaccuracies and/or uncertainties in its predictions outside the support of $\D$, and the policy optimization to maximize the preference (resp. reward), can exploit the such regions with spuriously high scores under $\pmle$ (resp. $\rmle$), resulting in a shift in the distribution of outputs  generated by the learned policies ($\pinash$ and $\pirlhf$). Colloquially, this phenomenon is often termed ``\emph{reward hacking}'' or ``\emph{preference hacking}''.
Pessimism in both the reward setting \citep{eisenstein2024helping, liu2024provably, fisch2024robust, cen2024valueincentivized} and the preference setting \citep{ye2024online} has been proposed as a way to remedy these issues. 

Pessimism in the reward setting leads to a $\max$-$\min$ objective,  
%\begin{align*}
    $\pipessrlhf \in \argmax_\pi \min_{r \in \R} \Jrlhf(\pi, r)$ %+ \lambda \Lreward(r,\D),
%\end{align*}
where $\R$ is an uncertainty set of reward functions, that is, all reward functions that are consistent with the dataset. \citet{liu2024provably} and \citet{fisch2024robust} show that for certain choices of $\R$, this game can be solved without actually maintaining the set $\R$ and performing the inner optimization in closed form. In the following section, we extend this to the preference setting (again, which includes reward-based BTL preferences), while also analyzing what forms of pessimism are most appropriate for learning reasonable optimal policies under uncertainty.\looseness=-1

\section{Pessimistic Preference-based RL}
We now define a natural extension of the pessimistic reward-based objective to the  case of preferences, and study its improvements and generalizations. Implementation issues in developing an efficient algorithm are deferred to \textsection\ref{sec:p3o}.

\paragraph{A pessimistic Nash solution.}
In preference-based RL, a pessimistic counterpart of the Nash solution in~\eqref{eq:nash} can be naturally formulated as
\begin{align}\label{eq:max-min-min-objective}
    \pipessnash & \in \argmax_\pi \min_{\pi'} \min_{p \in \cP} \Jpref(\pi, \pi', p)
\end{align}
where $\cP \subseteq \{\ \Y \times \Y \to [0,1] \}$ 
% where $\cP \subseteq [0,1]^{\Y\times\Y}$
defines an uncertainty set over preference functions.\footnote{The game not being symmetric leads to $\pipessnash \neq \pipessnash'$.} In particular we consider the  set $\cP(\D, c) \subset \{\Y\times\Y \to [0,1] \}$, for $c\geq 0$, which is defined as: 
\begin{align}\label{eq:uncertainity-set}
    \cP(\D, c) \doteq \{p : \Lpref(p;\D) \leq \Lpref(\pmle;\D) + c\},
\end{align}
and choose $\cP \doteq \cP(\D,c')$ for a value of $c'$, such that $p^\star \in \cP(\D,c')$. This formulation has been studied previously for certain choices of $\cP$ in the tabular \citep{cui2022when} and function approximation setting \citep{ye2024online, huang2024correcting}. These works prove that the solution $\pipessnash$ converges to the optimal policy if and only if a condition called \emph{unilateral coverage} holds, which requires that we can compare $\pipessnash$  with any response $y$, within the coverage of the sampling policy $\pisample$. That is, $p(\pipessnash, y)$ lies within a small interval as we vary $p\in\cP$, for all $y$. This approach has not been empirically evaluated in prior works, as the optimization problem is very challenging with no obvious practical strategies. Before discussing these algorithmic challenges, we will first explore the practical implications of the unilateral coverage requirement.\looseness=-1

\begin{figure}
    \centering
    \includegraphics[width=0.48\textwidth]{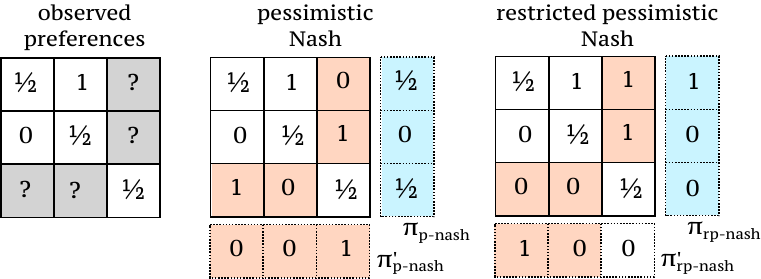}
    %\hspace*{1cm}
    \vspace{-12pt}
    \caption{
    An illustration of the problematic example for pessimistic preference optimization with unrestricted opponents. We assume that $\{y_1, y_2\}$ are well-sampled in the preference data, whereas \emph{$y_3$ is not}---resulting in certain preferences for $y_1$ vs. $y_2$, but completely uncertain preferences for $y_3$ vs. others (\textbf{Left}).
    %T $\cP$ that allows for all possible preferences against $y_3$.
    The $3\times3$ matrices above are then the optimized pessimistic preference matrices for $\{y_1, y_2, y_3\}$, with the $3d$ vectors the optimized competing policies.
    Specifically, shaded entries represent optimizable variables, and the values in each of the blue and red shaded entries are the solutions for the max ($\argmax_\pi$) and min player ($\argmin_{\pi'}$ and $\argmin_{p \in \cP}$), resp., see \eqref{eq:max-min-min-objective}. \textbf{Middle}: when the opponent $\pi'$ is unrestricted, the optimal policy $\pipessnash$  must hedge and put significant support on $y_3$. \textbf{Right}: restricting the support of $\pi'$ to the support of the \emph{preference-data} (i.e., $\{y_1, y_2\}$), avoids this issue, and yields a more reasonable optimal policy $\pirespessnash$.
    }
    \label{fig:bad_example}
    %\vspace{-10pt}
\end{figure}

\paragraph{The implications of an unconstrained min player.}
Consider an illustrative example in Figure~\ref{fig:bad_example} which is emblematic of typical RLHF scenarios. There is no context and $\Y = \{y_1, y_2, y_3\}$ and suppose further that we have that $p(y_1, y_2) = 1$ for all $p \in \cP$, so we are fully certain about this preference. But we never observe any comparisons involving $y_3$ in our preference data $(\pisample(y_3) = 0)$, and hence the set $\cP$ allows all values $p(y, y_3) \in [0,1]$ for $y \ne y_3$. To highlight the limitations of pessimism in preference optimization, we consider the problem in absence of regularization, i.e., $\beta = 0$. Then, as illustrated in Figure~\ref{fig:bad_example}~(Left) and proven in Appendix~\ref{sec:bad_example}, the pessimistic Nash policy $\pipessnash$ satisfies $\pipessnash(y_3) = 0.5$. That is, the policy takes an action completely out of the support of the sampled dataset w.p. $0.5$, where the ground-truth preferences can take completely arbitrary values. Intuitively, this happens because the pessimistic policy has to account for either possibility that $y_1$ is much better than $y_3$, or vice versa, since the set  $\cP$ is completely uncertain about the preference between $y_3$ and $y_1$.  Indeed, in most practical applications, many possible $y$'s will not be covered in the dataset, even distributionally, and it appears undesirable that the optimal policy obtained by pessimism will then \emph{predominantly} generate such outputs. 

This example highlights a key distinction between pessimism with rewards versus pessimism with preferences. When using pessimism with rewards, outputs which are not covered in the data tend to get a low score, and a reward-maximizing policy naturally avoids such outputs. But in preference-based learning, the min player $\pi'$ can choose any output which is not observed in the preference data, and pessimism over preferences means that the max player looses to this action in the worst case---forcing the max player to put probability over such an uncovered output as well. We now propose a remedy for this issue.
% \begin{wrapfigure}{r}{0.4\textwidth}

\paragraph{Restricted pessimistic Nash with a constrained min player.} 
Given the example from Figure~\ref{fig:bad_example}, an intuitive response is to consider a Nash strategy where the support for the opponent policy $\pi'$ is restricted to actions which are well-sampled in the preference dataset. For tabular scenarios, where we have no contexts and a finite $\Y$, with all possible policies in the class $\Pi$, such a restriction can be carried out by explicitly constraining the support of the min-player in the pessimistic objective~\eqref{eq:max-min-min-objective}. However, this does not generalize to more practical scenarios with parametric policies over a large output space. Instead, for such situations, we  define a subset $\Pi(\pisample, C) \subseteq \Pi$ to be a set of policies whose outputs are ``well-covered'' by the sampling policy $\pisample$, with $C$ denoting a coverage parameter that we define below.\looseness=-1 

\begin{definition}[Covered policy set] For a given sampling policy $\pisample \in \Pi$ and constant $C$, the covered policy set $\Pi(\pisample, C)$ with respect to $\pisample$ is the set of policies  such that $\forall \pi,\pi' \in  \Pi(\pisample, C)$ and $\forall p_1,p_2 \in \cP$,
% for any $\pi, \pi' \in \Pi(\pisample, C)$ and any pair of preference functions $p_1, p_2 \in \cP$ , we have:
\begin{align}
\begin{split}
    &\E_{y \sim \pi, y' \sim\pi'} (p_1(y, y') - p_2(y, y'))^2  \\ &\hspace{0.5cm}\leq C\cdot \E_{ y, y' \stackrel{\mathrm{i.i.d.}}{\sim}\pisample} (p_1(y, y') - p_2(y, y'))^2.
    \label{eq:support}
\end{split}
\end{align}
\label{def:covered-policy-set}
\end{definition}

What covered policies are included in $\Pi(\pisample, C)$?  Clearly, $\pisample \in \Pi(\pisample, C)$ when $C \geq 1$. We assume $C \geq 1$ in the sequel to ensure that this containment happens. We can also show (see Appendix~\ref{app:bounded_likelihood_coverage} for a short derivation) that $\Pi(\pisample, C)$ includes the set of all policies with likelihood ratios with respect to $\pisample$ that are uniformly bounded by $\sqrt{C}$, that is, $\Pi_{\sqrt{C}} \subseteq \Pi(\pisample, C)$, where $\Pi_{\sqrt{C}} = \{\pi\in\Pi~:~\|\pi/\pisample\|_\infty \leq \sqrt{C}\}$. %\footnote{See Appendix~\ref{app:bounded_likelihood_coverage} for a short derivation.}

In fact, we further show in Appendix~\ref{app:covariance_linear} that when the preference functions are linear in a shared feature map, then this coverage condition is ensured whenever the cross-covariance matrices of $\pi$ and $\pi'$ are  sufficiently aligned with the covariance matrix of $\pisample$.

Perhaps most importantly, however, when defining the following restricted pessimistic Nash solution,  $\pirespessnash$, using this notion of coverage, i.e.,
\begin{equation}
    % \hat \pi 
    \pirespessnash \in \argmax_{\pi \in \Pi} \min_{\pi'\in\Pi(\pisample, C)} \min_{p\in\cP} \Jpref(\pi, \pi', p),
    \label{eq:restricted-pess-nash}
\end{equation}
we can give the following performance guarantee for $\pirespessnash$ under the \emph{ground-truth preference function} $p^\star$ which generated the preference dataset $\D$  (which was then used to derive the set of plausible preference functions $\cP$).

\begin{lemma}[Preference guarantee for the restricted pessimistic Nash policy]
    We denote the restricted pessimistic Nash policy by $\pirespessnash$  from \eqref{eq:restricted-pess-nash}, and let $p^\star$ be the ground-truth preference function underlying $\D$. Then we have that for any $\pi\in\Pi(\pisample, C)$ with $C \geq 1$:
    \begin{equation*}
        p^\star(\pirespessnash, \pi) \geq \frac{1}{2} - 2\sqrt{C\epsilon},
    \end{equation*}
    where $\epsilon$ is a bound on how much preference functions in $\cP$ can disagree in total variation under $\pisample$: $\E_{( y, y')\stackrel{\mathrm{i.i.d.}}{\sim} \pisample} |p_1(y, y') - p_2(y, y')| \leq \epsilon$, $\forall p_1, p_2 \in \cP$.
\label{lem:restricted-nash}
\end{lemma}
Appendix~\ref{appendix:lem:restricted-nash-proof} restates Definition~\ref{def:covered-policy-set} and Lemma~\ref{lem:restricted-nash} with the context $x$ included, along with a proof.
This result shows that the restricted pessimistic Nash policy is always preferred to all other covered policies in $\Pi(\pisample, C)$ up to an error term. In comparison, the unrestricted pessimistic Nash solution from \eqref{eq:max-min-min-objective} does not satisfy this guarantee in general. To see this, consider the example in Figure~\ref{fig:bad_example}, where the unrestricted pessimistic Nash policy $\pipessnash = [1/2, 0, 1/2]$ is dispreferred to the covered policy $\pi_1 = [1, 0, 0]$ with probability $p^\star(\pi_1, \pipessnash) = 1/2 + \gamma/2$ when $p^\star(y_1, y_3) = \gamma$.
 
As this example suggests, the unrestricted pessimistic solution can be arbitrarily dispreferred, even to covered policies. Existing analyses by  \citet{cui2022when} for the unrestricted version indeed do not apply to this example since $\pisample(y_3) = 0$ and, thus, the  necessary unilateral coverage condition is violated \cite{cui2022when}. In fact, no policy can satisfy unilateral coverage in this case, and we get a vacuous guarantee out of their analysis. The same is true for the relaxed coverage condition of \citet[][Appendix B]{zhang2023offline}, as it still applies to the unrestricted pessimistic solution.\looseness=-1

In some sense, the contrast between our result from Lemma~\ref{lem:restricted-nash} and those of \citet{cui2022when} is analogous to the classical analysis of offline RL methods (see, e.g.,~\citet{chen2019information}) and pessimistic offline RL techniques~\citep{xie2021bellman}. Without pessimism in offline RL, we end up with vacuous guarantees, while the pessimistic results allow a non-trivial sub-optimality bound against any policy well covered by the data collection policy. Similarly, the results of \citet{cui2022when} offer a strong guarantee when the data collection policy is sufficiently exploratory, but are rendered vacuous without this. In contrast, our analysis of the restricted Nash estimator offers an opportunistic guarantee, where we are able to adaptively compete with all policies which are well covered by $\pisample$.
We note that these considerations are particularly pertinent when aligning LLMs with small preference datasets, where the output space is of long sequences over a large vocabulary, of which the alignment data typically only covers a tiny sliver, leaving no hope for unilateral coverage style assumptions to be satisfied.\looseness=-1

\section{P3O: An Efficient Implementation}
\label{sec:p3o}

\begin{algorithm}[tb]
   %\alglong (\alg($\alpha$))}
   \label{alg:pspo}
\begin{algorithmic}
   \STATE {\bfseries Hyperparameters:} Mixing coefficient $\alpha$,  regularization coefficient $\beta$, preference regularization coefficient $\lambda$, EMA parameter $\gamma$, learning rates $(\eta_p, \eta_\pi)$
   \STATE {\bfseries Initialize:} $\bar{\pi}_1 = \pi_1 = \piref$, $p_1 = \pmle$
   \FOR{$t=1, 2, \dots$}
      \STATE Set $\pimixalpha \propto \bar{\pi}^{1-\alpha}_t \pisample^{\alpha}$ as mix of $\piref$ and the exponential moving average $\bar{\pi}_t$ for restricted Nash. 
      \STATE Approximate the current objective \eqref{eq:p3o}:
      \begin{align*}
          \Jpppo{\alpha}(\pi_t, p_t) \doteq 
          p_t(\pi_t, \pimixalpha) &- \beta \KL{\pi_t}{\piref} \\
          & - \lambda \mathrm{KL}_{\piref}\left(\pmle\|p\right)
        %   \EE{y,y'\sim \piref}{\KL{\pmle(y,y')}{p(y,y')}}
      \end{align*}
      \STATE Update $\pi_{t+1} \leftarrow \pi_t + \eta_\pi \frac{\partial \Jpppo{\alpha}(\pi, p_t)}{\partial \pi} \bigg|_{\pi = \pi_t}$
      \STATE Update $p_{t+1} \leftarrow p_t - \eta_p \frac{\partial \Jpppo{\alpha}(\pi_t, p)}{\partial p} \bigg|_{p = p_t}$
      \STATE Update $\bar{\pi}_{t+1} \leftarrow \gamma \pi_t + (1-\gamma) \bar{\pi}_t$
   \ENDFOR
\end{algorithmic}
\caption{\alg\!($\alpha$)}
\end{algorithm}
% \vspace{-10pt}
%\end{figure}

We now develop an efficient algorithm,   \alglong~(\alg), that approximately solves the restricted pessimistic Nash formulation.\looseness=-1

\paragraph{Approximating the restricted policy set.} A first obstacle to an efficient algorithm is that the definition $\Pi(\pisample, C)$ is not amenable to easy implementation. However, in KL-regularized preference-based RLHF, there is a natural heuristic to approximate this restriction via an additional KL regularization term. Recall that the central goal of $\Pi(\pisample, C)$ is to limit optimization to policies $\pi$ which generate responses that are in the coverage of the data generating policy $\pisample$. We encourage this through adding an additional penalty based on the KL divergence between $\pi'$ and $\pisample$:
\begin{align}\label{eq:restricted-objective}
    &\max_\pi \min_{p \in \cP} \min_{\pi'} \;\; p(\pi, \pi') - \beta \KL{\pi}{\piref} \\ \nonumber
    &\quad + (1-\alpha) \beta \KL{\pi'}{\piref} + \alpha\beta \KL{\pi'}{\pisample}.
 \end{align}
Note that $\pi'$ is regularized with respect to both the reference policy $\piref$ and the sampling policy $\pisample$, where the added parameter $\alpha$ controls the relative strength of the contribution of $\piref$ versus $\pisample$. While going from a data-aware constraint in terms of $\Pi(\pisample, C)$ to KL regularization is lossy, we note that this is for the $\min$ player $\pi'$ and only affects the max player $\pi$ through the data-dependent $p$ term.\looseness=-1

We use \alg\!($\alpha$) to denote this variant, with \alg being the shorthand for \alg\!(0).
Using a closed-form solution to the inner KL-regularized problem for $\pi'$, we next show how to obtain an equivalent, but greatly simplified, objective for $\pi$. 
First, we define the shorthand $\pimixalpha(y; \pi_1, \pi_2)$ as:
\begin{align}\label{eq:pi-mix}
\begin{split}
    \pimixalpha(y; \pi_1, \pi_2) \propto  \pi_{1}^{1-\alpha}(y)\pi_2^{\alpha}(y).
\end{split}
\end{align}
Optimizing $\pi$ against an appropriate mixed distribution is then equivalent to solving for $\eqref{eq:restricted-objective}$, as we show below.

\begin{lemma}
  The optimization problem in \eqref{eq:restricted-objective} is equivalent to the following objective, assuming that the minimization over $\pi'$ is over all possible policies:
%  $\Delta(\Y)$,
\begin{align}\label{eq:restricted-objective-logsumexp}
\begin{split}
     \max_\pi \min_{p\in\cP} & - \log \E_{y\sim\pimixalpha(\piref, \pisample)} \left[e^{ \frac{-p(\pi, y)}{\beta}}\right] 
     \\
     & \quad - \beta \KL{\pi}{\piref}.
\end{split}
\end{align}
 \label{lem:exact-obj}
 \vspace{-5pt}
\end{lemma}

We provide a proof of this equivalence in Appendix~\ref{appendix:nash-restricted-proof}. We also note that replacing $\pimixalpha(\piref, \pisample)$ with $\piref$ (i.e., $\alpha = 0$) gives an equivalent rewriting for the pessimistic Nash objective with no support restrictions, as given in \eqref{eq:max-min-min-objective}.

\paragraph{Approximating the log-partition function.}
The objective in Lemma~\ref{lem:exact-obj} simplifies the inner minimization to only have one variable, but at the cost of changing the objective to have a more complicated log-partition function term. Consequently, we can no longer get unbiased stochastic gradients of the objective from a mini-batch of data, due to the non-linearity of the logarithm outside of the expectation. 

To obtain a practical algorithm, we leverage ideas from variational inference~\citep{jordan1999introduction} to approximate the log-partition function. Doing so, we obtain the following result, which is proved in Appendix~\ref{appendix:approx-obj-proof}.

\begin{lemma}
For any choice of policies $\pi, \bar\pi \in \Pi $:
\begin{align}
\label{eq:upper_bound}
    &\min_{p\in\cP} - \log \E_{y\sim\pimixalpha(\piref, \pisample)} \left[e^\frac{-p(\pi, y)}{\beta}\right] \nonumber \\
    % &\hspace{0.2cm}= \min_{p\in\cP} -\log \E_{y\sim\pimixalpha(\bar\pi, \pisample)} \left[\left(\frac{\piref(y)}{\bar\pi(y)}\right)^{1-\alpha}e^\frac{-p(\pi, y)}{\beta}\right] \nonumber \\
    &\hspace{0.75cm}\leq \min_{p\in\cP} \E_{y\sim\pimixalpha(\bar\pi, \pisample)} \left[\frac{p(\pi, y)}{\beta}\right] + \kappa,
\end{align}
where $\kappa$ is independent of $\pi$ and $p$. %Furthermore, equality holds when $\bar\pi \propto \pimixalpha(\piref, \pisample) \exp(-p(\pi, y) / \beta)$.
\label{lem:approx-obj}
\end{lemma}

Due to the direction of the inequality, Lemma~\ref{lem:approx-obj} gives only an \emph{upper bound} for our objective in Lemma~\ref{lem:exact-obj}, and therefore maximizing the two is not equivalent. Nevertheless, the approximate objective is tractable, and simply takes the form of optimizing preferences against some comparator $\bar\pi$ mixed with the sampling distribution. Furthermore, the approximate objective at any fixed value of $p$ is tight when $\bar\pi(y)^{1-\alpha} \propto \exp(-p(\pi, y) / \beta)$, which resembles the multiplicative weight updates observed in prior self-play algorithms~\citep{swamy2024minimaximalist, munos2023nash}.
Since the current preference function iterate $p_t$ is slowly moving during gradient descent, with this motivation in hand,  we choose the competitor policy $\bar\pi$ to be an exponentially moving average of past policy iterates in our experiments, giving our algorithm a pessimistic self-play flavor.

\paragraph{Approximating the preference uncertainty set.}
As a final step, we replace the constrained optimization over the preference uncertainty set $\cP$ with an unconstrained optimization over all preference functions in some parametric family by adding an additional loss term $-\Lpref(p; \D)~$\citep{liu2024provably}, corresponding to the Lagrangian form of the constraint defining $\cP$.  The objective function then becomes 
\begin{align*}
% \label{eq:p3o-ll}
    \E_{y\sim\pimixalpha(\bar\pi, \pisample)} [p(\pi, y) ] - \beta \KL{\pi}{\piref} - \lambda \Lpref(p; \D),
\end{align*}
where we rescaled the objective to absorb the $1/\beta$ on $p$ into the corresponding hyper-parameters of the KL and likelihood loss components (i.e., $\beta, \lambda$). The above objective requires us to also load the preference dataset $\D$ while trying to learn the policy, which can be somewhat inconvenient. To circumvent this issue, however, we can instead simply regularize the preference model to stay close to $\pmle$, i.e.,
\begin{align}\label{eq:p3o}
    \Jpppo{\alpha}(\pi, p) \doteq&~ 
    %  \E_{y\sim\pimixalpha(\bar\pi, \pisample)} \left[p(\pi, y) \right] 
    % \\ &- \beta \KL{\pi}{\piref}  - \lambda \mathrm{KL}_{\pisample}\left(\pmle\|p\right),\nonumber
    p\Big(\pi, \pimixalpha(\bar\pi, \pisample)\Big) 
    \\ &- \beta \KL{\pi}{\piref}  - \lambda \mathrm{KL}_{\pisample}\left(\pmle\|p\right),\nonumber
\end{align}
where $\mathrm{KL}_{\pisample}\left(\pmle\|p\right)$ is defined as
\begin{align*}
\EE{y,y'\stackrel{\mathrm{i.i.d.}}{\sim} \pisample}{\KL{\pmle(y,y')}{p(y,y')}}.
\end{align*}
% $\KL{\pmle}{p} \doteq \EE{ y,y'\sim \pisample }{ \pmle(y,y') \ln \tfrac{\pmle(y,y')}{ p(y,y')} + (1 - \pmle(y',y))\ln \tfrac{1-\pmle(y',y)}{ 1-p(y',y)}  }$
If the MLE solution $\pmle$ is a good approximation to the true preferences $p^\star$ on $y, y' \stackrel{\mathrm{i.i.d.}}{\sim} \pisample$, then this KL divergence provides a good approximation to the likelihood-based version.
The resulting algorithm is shown in Algorithm~\ref{alg:pspo}. 

Having introduced \alg, which handles general preferences, we can  extend it to the special case where preferences are parameterized by a reward function under the BTL model. In this setting, we replace the general preference function $p$ with $\pbt(r)$ in \eqref{eq:p3o}, resulting in the following objective:
\begin{align}\label{eq:prpo}
    &\Jprpo{\alpha}(\pi, r) \doteq
     \pbt\Big(\pi, \pimixalpha(\bar\pi, \pisample){;}r\Big)  \\ &\quad-\beta \KL{\pi}{\piref} 
     - \lambda \mathrm{KL}_{\pisample}\left(\pbt(\rmle)\| \pbt(r)\right). \nonumber
\end{align}
We refer to the algorithm that optimizes \eqref{eq:prpo} as \prpolong (\prpo). Appendix~\ref{appendix:prpo} provides the pseudo code and further discussion on \prpo.

\section{Experimental Results}
\label{sec:experiments}
\begin{figure}
    \centering
    \includegraphics[width=0.49\textwidth]{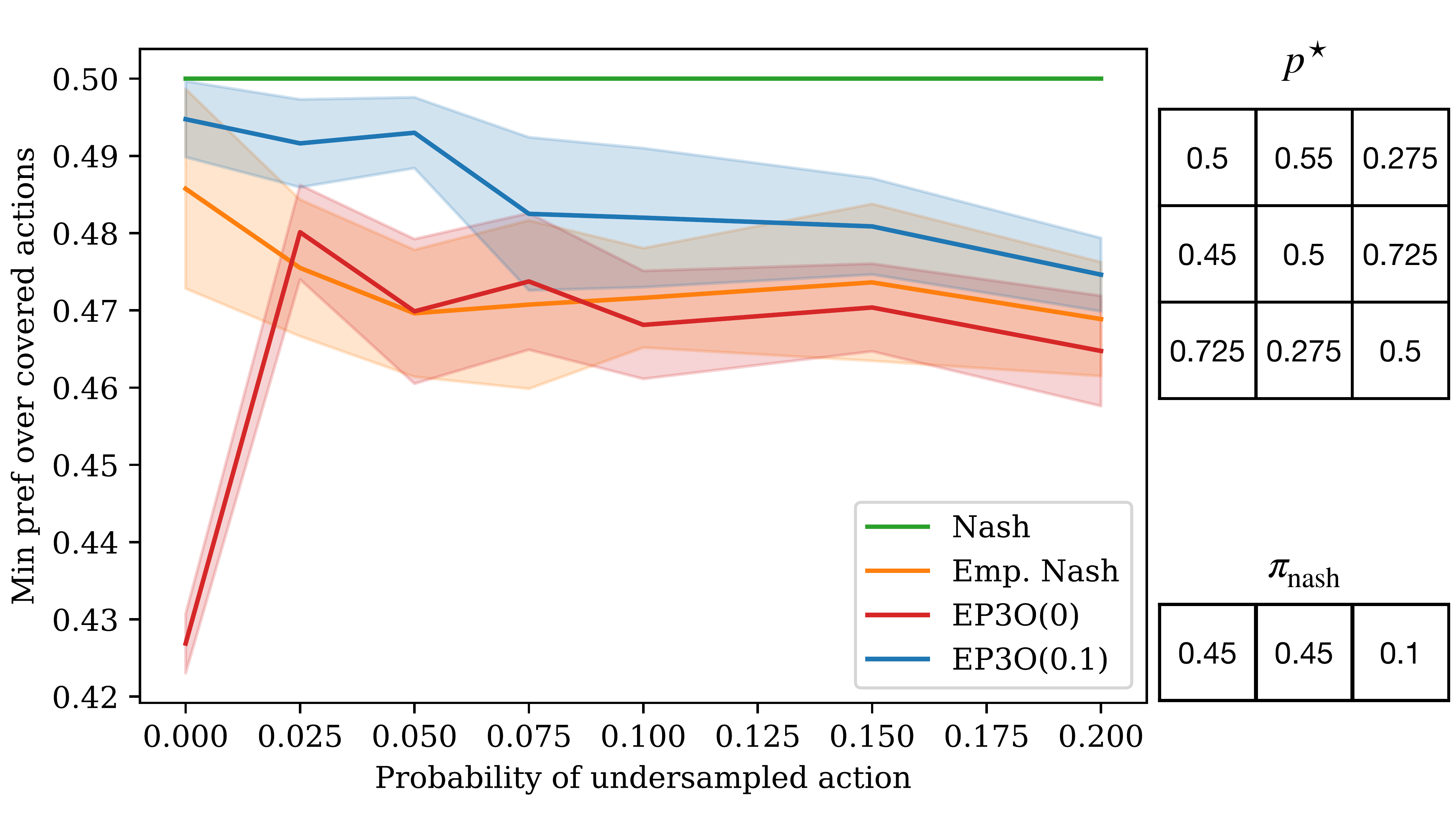}
    \vspace{-12pt}
    \caption{
    \textbf{Tabular experiments:}
    Comparison of the different objectives with an explicit search over the main policy, opponent policy, and version space.
    The $X$-axis shows the probability assigned to the under-sampled output ($y_3$), from $0.0 \rightarrow 0.2$.
    The $Y$-axis indicates the minimum preference of the policy found over all covered actions (higher is better).
    We also show the ground-truth preference matrix $p^\star$ (top-right) and the corresponding $\pinash$ (bottom-right). Results are averaged over 10 random seeds, shaded areas represent $\pm 2 \times$ std error. EP3O(0.1) corresponding to the restricted Nash formulation consistently does well, particularly when the sampling rate of $y_3$ is very low (left part of plot).\looseness=-1
    }
   % \vspace{-10pt}
    \label{fig:tabular-experiments}
\end{figure}

In this section, we illustrate through multiple experiments the effectiveness of the \alg approach. We begin with evaluating the solution $\pirespessnash$ to the exact objective \eqref{eq:restricted-objective-logsumexp} by performing a brute force search over all the variables in a small tabular environment to understand how the objective performs for simple games. 
Next, we perform experiments on summarization and helpfulness tasks using the approximate objective $\Jpppo{\alpha}$~\eqref{eq:p3o} to demonstrate its effectiveness. 

\begin{figure*}[!ht]
    \centering
    \includegraphics[height=1.35in]{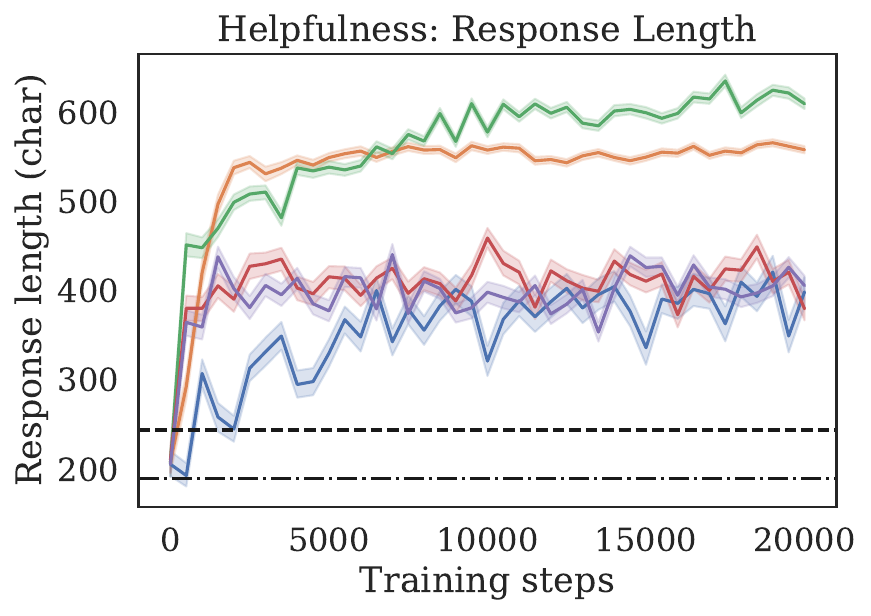}
    \includegraphics[height=1.35in]{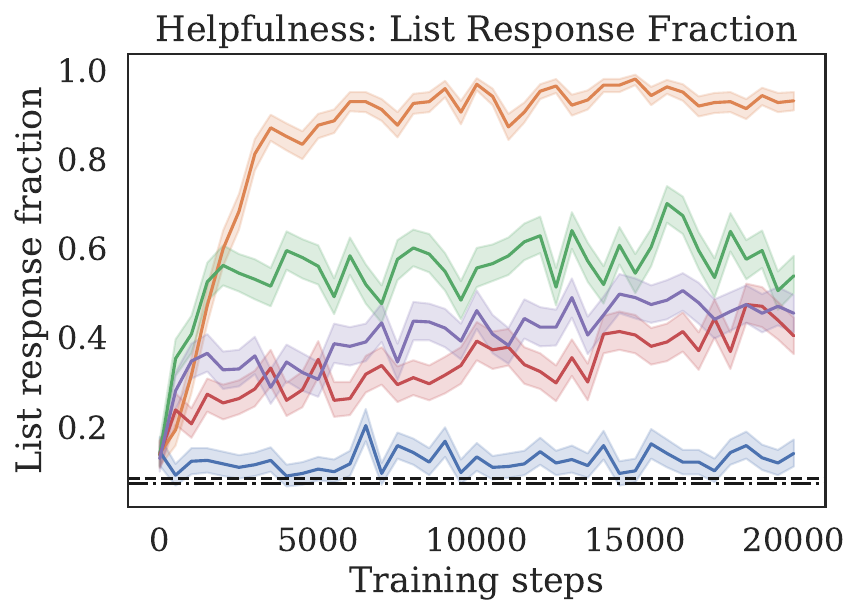}
    \includegraphics[height=1.35in]{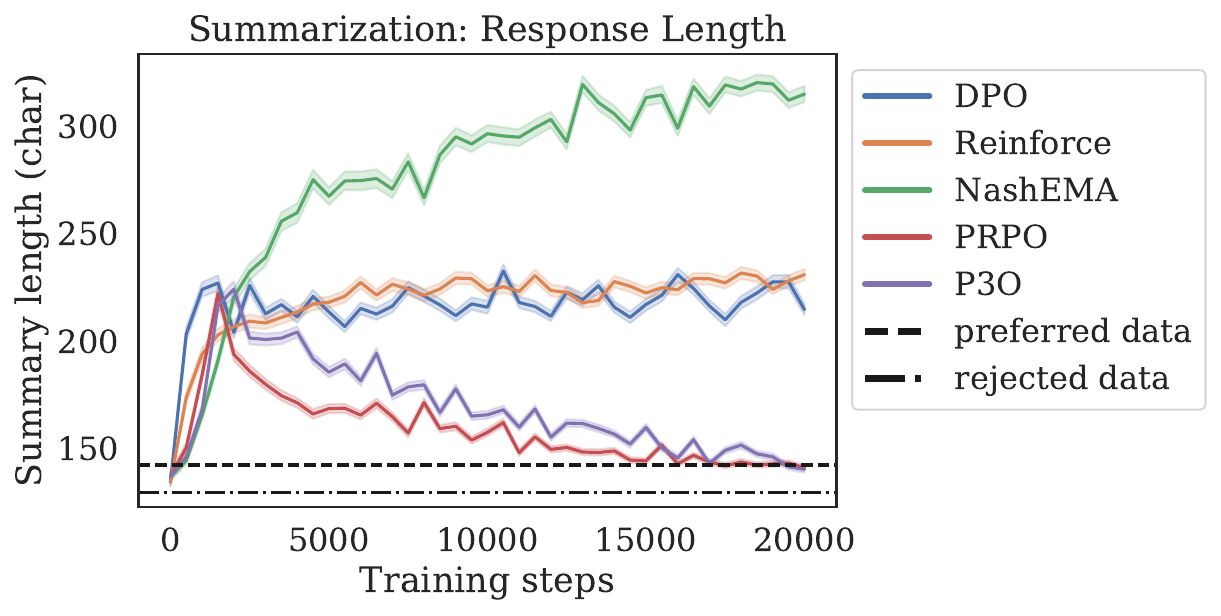}
    \vspace{-7pt}
    \caption{\textbf{Qualitative results for the  helpfulness and summarization tasks:} 
      On helpfulness, both response length and list formats are common reward hacks~\cite{eisenstein2024helping}. While policies do generate longer responses than the responses in $\pisample$, both REINFORCE and Nash-EMA converge on generations that are  $\approx 40-50\%$ longer than those of P3O and PRPO (\textbf{left}).  Similarly, REINFORCE degenerates into producing responses that are nearly all formatted as lists (with Nash-EMA at over $50\%$), while P3O and PRPO stay closer to $\pisample$ (\textbf{middle}). On summarization,  DPO, REINFORCE, and Nash-EMA all show clear signs of length hacking---also a pervasive issue on this task~\cite{eisenstein2024helping, singhal2023long, park2024disentangling}. In contrast, both PRPO and P3O converge to the average length of preferred responses in $\pisample$, all while also achieving the highest win-rates (\textbf{right}).}\looseness=-1

    \label{fig:qualitative-results}
    % \end{minipage}
\end{figure*}

\begin{figure}
    \centering
    \includegraphics[height=1.32in]{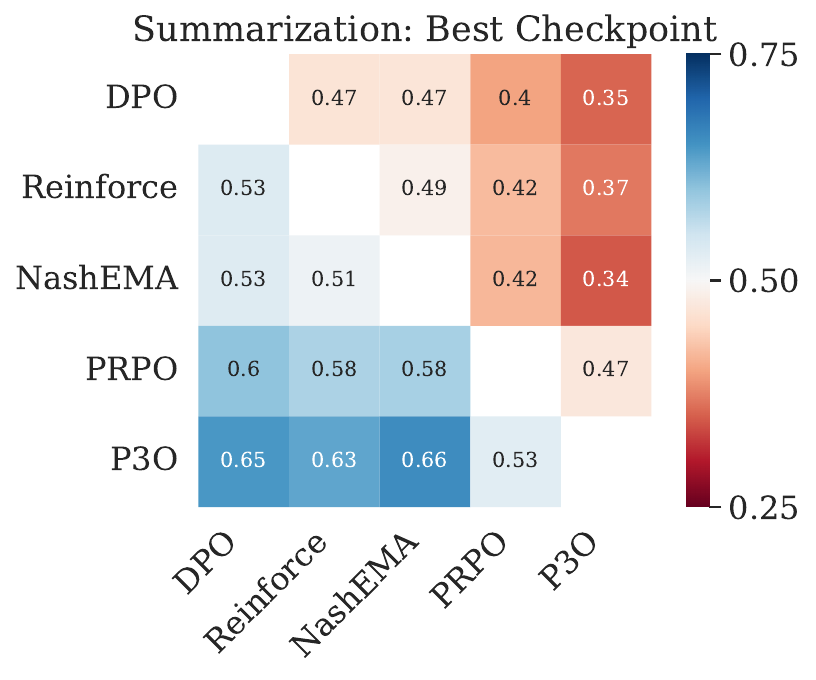}
    \includegraphics[height=1.32in]{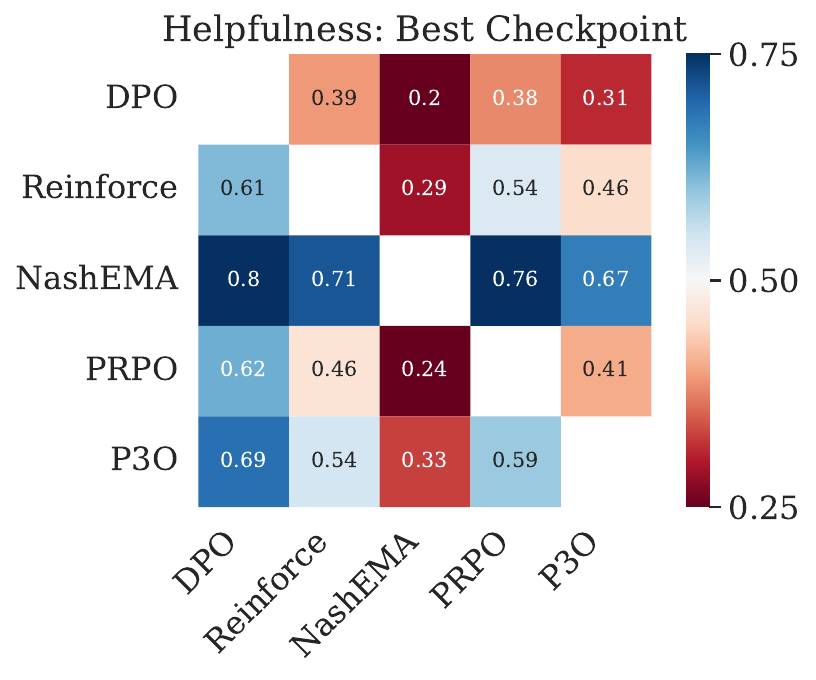}
    \vspace{-7pt}
    \caption{Confusion matrix showing the pairwise evaluation across the different methods, for each method's best checkpoint selected in terms of preferences over the reference policy. In head-to-head comparisons, \alg and \prpo  outperform baselines, except Nash-EMA on helpfulness, where the Gemini 1.5 Flash evaluation preference aligns with the preference model used for training---favoring the verbose, list-heavy outputs of Nash-EMA (and REINFORCE, to a lesser degree). However, when looking for both  helpful  \emph{and} concise responses, P3O and PRPO win rates increase dramatically against Nash-EMA (see Figure~\ref{fig:confusion-concise} in the appendix), demonstrating their robustness.}
    
    \label{fig:confusion-best}
\end{figure}
% \subsection{Tabular Experiments}
\paragraph{Tabular experiments.}
We first conduct experiments in a controlled, tabular setting with three possible outputs $y \in \{y_1, y_2, y_3\}$,  and a ground-truth preference matrix \(p^\star\). We vary the probability of sampling $y_3$ from $0.0$ to $0.2$, distributing the remaining probability equally among $(y_1, y_2)$. Thus, $y_3$ is consistently the under-sampled output. We conduct experiments with varying Nash strategies, including cases where $y_3$ is preferred, as well as dispreferred over the other two actions. Complete details about the experimental setup and search procedure are provided in Appendix~\ref{appendix:tabular-experiments}.

Figure~\ref{fig:tabular-experiments} illustrates a case in which the under-sampled action is dispreferred ($\pinash(y_3)=0.1$). We compare Exact \alg\!(0.1) (E\alg\!(0.1)), E\alg\!(0), as defined in Eq.~\eqref{eq:restricted-objective-logsumexp}, and the non-pessimistic Nash policy obtained from $\pmle$. 
Additional scenarios, including ones where the under-sampled action is genuinely preferred, appear in Figure~\ref{appendix:fig:tabular-experiments} (see Appendix~\ref{appendix:tabular-experiments}).
Notably, E\alg\!(0.1) serves as a robust default choice: when the under-sampled action is truly dispreferred (particularly under extremely low sampling), the restricted pessimism in E\alg\!(0.1) prevents the policy from overcommitting to an insufficiently explored but suboptimal action. 
Conversely, if the under-sampled action is actually favored under the true $\pinash$, even a small amount of data may guide non-pessimistic methods to weight that action correctly—potentially yielding strong performance. 
Since we typically lack ground-truth preferences, E\alg\!(0.1) offers a ``safe-default'' strategy.\looseness=-1

\paragraph{Task 1: Summarization.} To demonstrate the effectiveness of our approach in mitigating preference hacking,  we compare it against existing preference optimization methods on the popular TL;DR summarization task \citep{volske2017tl, stiennon2020learning}. Following prior studies on reward hacking for TL;DR \citep{eisenstein2024helping}, we train an MLE preference model $\pmle$ as well as MLE reward model $\rmle$ by fine-tuning a T5 XL (3B) model \citep{raffel2020exploring, roberts2023scaling} on the preference dataset. The initial policy $\piref$ is obtained by supervised fine-tuning  a T5 large model (770M) on the human reference summaries in TL;DR. Choosing a larger preference model than the policy is a commonly employed strategy for mitigating hacking~\citep{eisenstein2024helping}. We initialize the training preference model $p_1 \doteq \pmle$ in case of \alg, and $r_1 \doteq \rmle$ in case of \prpo.

\paragraph{Task 2: Helpfulness.} We also test our method at larger scales on the Anthropic Helpfulness task~\citep{bai2022training} using 8B PaLM-based~\citep{anil2023palm} policy, reward, and preference models. The helpfulness tasks consists of dialogues between humans and an automated assistant. The goal is to complete the next turn of the assistant by producing an engaging and helpful response. Like before, we obtain MLE preference and reward models by fine-tuning a pre-trained PaLM model on the preference data. These checkpoints are also used to initialize the  preference and reward models for P3O and PRPO. The initial policy $\piref$ is obtained from an instruction-tuned PaLM model.

We compare \textbf{\alg} and \textbf{\prpo} against existing popular non-pessimistic RLHF methods,  \textbf{\dpo}~\citep{rafailov2023direct} and \textbf{REINFORCE}~\citep{willams1992simple} (which has been shown to outperform PPO~\citep{ahmadian2024back}), as well as the preference-model-based method \textbf{Nash-EMA}~\cite{munos2023nash} (which is also non-pessimistic).
We also compare to \textbf{PDPO}~\citep{fisch2024robust}, a pessimistic, offline variant of DPO, in Appendix~\ref{app:experiments}. 
%In Appendix~\ref{app:experiments}, we further show a comparison with a preference-model based, non-pessimistic baseline.
We set $\alpha=0$ for our methods, as in these experiments $\piref$ and $\pisample$ are similar, and hence $\alpha$ does not influence the behavior of \alg.

All policies are evaluated using preferences assigned by Gemini~1.5~Flash \citep{geminiteam2024geminifamilyhighlycapable}.\looseness=-1

\paragraph{Results.} Figure~\ref{fig:eval_pref_gemini} shows that the non-pessimistic baselines plateau at a lower overall performance level in general, with \dpo even degrading substantially with prolonged training. In contrast, both \alg and \prpo achieve a significantly higher preference over \piref, surpassing the baselines (a result which is further supported by the confusion matrices in Figure~\ref{fig:confusion-best}). The lone exception is Nash-EMA on helpfulness, which performs very strongly according to our Gemini 1.5 evaluator---a result that happens to be helped by the alignment of the Gemini 1.5 evaluation preference with Nash-EMA's generally longer  responses. Under an evaluation that emphasizes both helpful \emph{and} concise responses, \alg and \prpo are significantly better (Figure~\ref{fig:confusion-concise}). Notably, the preference-based \alg outperforms the reward-based \prpo, indicating the advantage of general preference models.\looseness=-1

Figure~\ref{fig:qualitative-results} also illustrates how pessimism results in qualitatively different outputs by helping to avoid some of the common ``hacking'' behaviors related to length and style across different tasks that have been well-reported in the literature~\cite{eisenstein2024helping, singhal2023long, park2024disentangling}. %In the left panel, \dpo quickly diverges from \piref, while \alg, \prpo\xspace with $\beta=10^{-5}$, and REINFORCE with $\beta=0.1$ remain closer to $\piref$. 
In particular, a typical form of reward hacking in summarization is ``length hacking'', where policies inflate their rewards by producing overly verbose responses. % In helpfulness, prior work has also found that policies tend to produce responses that are predominantly composed in list formats. 
We observe that non-pessimistic methods generate summaries significantly longer than those in the original dataset. In contrast, \alg and \prpo converge to summary lengths that are more consistent with $\piref$. Interestingly, while REINFORCE  exhibits length hacking whereas \alg and \prpo do not, in Figure~\ref{fig:kl-plots} (left) we see that REINFORCE actually maintains a lower KL divergence from $\piref$ compared to \alg\xspace and \prpo. This suggests that \alg and \prpo are learning  policies that are substantially distributionally distant from $\piref$, but along quality dimensions that are distinct from simply length (and that result in higher perceived quality in our evaluations in Figure~\ref{fig:eval_pref_gemini}). Additionally, for helpfulness, we observe that REINFORCE tends to format nearly all responses as lists, much higher than the prevalence of lists in the preference data (Figure~\ref{fig:qualitative-results}, middle), an artifact not shared by the responses of \alg or \prpo. We also observe a significant inflation in response lengths for both REINFORCE and Nash-EMA in helpfulness, converging to nearly 1.5x the response lengths for \alg and \prpo. In Appendix~\ref{app:experiments}, we present further results on the training dynamics which indicate a steadily improved objective for all the methods, even when the eval performance is non-monotonic. We also show some cherry-picked responses for both summarization and helpfulness tasks to illustrate the stylistic differences in their responses. We refer the reader to Appendix~\ref{app:experiments} for these and further details and results of our empirical evaluation.\looseness=-1

\section{Conclusion}
Modern RLHF methods suffer from a significant tendency to overoptimize spurious preferences (or rewards) that are derived from faulty preference (or reward) models. In this work, we introduced pessimistic, preference-based RLHF objectives, which carefully balance uncertainty with effective learning.  In particular, we theoretically analyzed the limitations of existing pessimistic estimators, and derive a novel formulation for a restricted, pessimistic Nash solution with provable advantages. Empirical results on multiple tasks and models demonstrate that our approach effectively resists overoptimization while outperforming standard RLHF baselines---highlighting the potential of pessimistic objectives for achieving robust language model alignment.\looseness=-1

\section*{Impact Statement}

This paper introduces new ideas to the active field of research on large language model post-training, which we hope will help  facilitate the successful alignment of models to human preferences, as well as  improve our understanding of current approaches---with the goal of ultimately supporting the development of capable and  reliable AI systems that are easier, and more stable, to finetune.

\bibliography{references}       % Points to the BibTeX file (reference.bib)
\bibliographystyle{icml2025}

%%%%%%%%%%%%%%%%%%%%%%%%%%%%%%%%%%%%%%%%%%%%%%%%%%%%%%%%%%%%%%%%%%%%%%%%%%%%%%%
%%%%%%%%%%%%%%%%%%%%%%%%%%%%%%%%%%%%%%%%%%%%%%%%%%%%%%%%%%%%%%%%%%%%%%%%%%%%%%%
% APPENDIX
%%%%%%%%%%%%%%%%%%%%%%%%%%%%%%%%%%%%%%%%%%%%%%%%%%%%%%%%%%%%%%%%%%%%%%%%%%%%%%%
%%%%%%%%%%%%%%%%%%%%%%%%%%%%%%%%%%%%%%%%%%%%%%%%%%%%%%%%%%%%%%%%%%%%%%%%%%%%%%%
\newpage
\appendix
\onecolumn

\section{Extended Literature Review}
% --- Offline RL and Pessimism ---
Offline RL is primarily concerned with learning a policy from a fixed dataset, a problem that has attracted considerable attention. Many works focus on scenarios with sufficiently broad dataset coverage~\citep{antos2007fitted, antos2008learning, munos2003error, munos2008finite, farahmand2010error, chen2019information, xie2020q}, though such assumptions tend to be overly restrictive and seldom hold in real-world situations. Consequently, recent research has shifted toward the more realistic setting of inadequate coverage~\citep{wang2020what}, aiming to learn a ``best effort'' policy~\citep{liu2020provably}. Two major strategies have emerged to handle poor coverage: behavior policy regularization~\citep{fujimoto2019off, laroche2019safe, kumar2019stabilizing, wu2019behavior, jaques2019way, kostrikov2021offline, xiao2023sample} and pessimism in the face of uncertainty~\citep{kumar2020conservative, liu2020provably, kidambi2020morel, yu2020mopo, buckman2020importance, jin2021pessimism, zanette2021provable, xie2021bellman, cheng2022adversarially, zhang2024pessimism, koppel2024information}. In limited-data regimes, \emph{pessimism} has been shown to provide strong theoretical guarantees for the resulting policy~\citep{buckman2020importance, jin2021pessimism}, achieving min-max optimality in linear MDPs. Moreover, it has been successfully incorporated into both linear~\citep{zanette2021provable} and deep RL (DRL) settings~\citep{xie2021bellman, cheng2022adversarially}.

% --- Pessimism in RLHF ---
Behavior policy regularization has also been explored in language models~\citep{jaques2019way}, alongside standard RLHF approaches that commonly regularize to a reference policy~\citep{stiennon2020learning}. When the reward function is learned from limited data, inaccuracies naturally arise, mirroring the challenge in value-based offline RL where value estimates become unreliable in underrepresented state-action regions. Pessimism thus serves as a compelling remedy and has recently been investigated in the standard reward-based setting. Concurrently, several works~\citep{fisch2024robust, liu2024provably, cen2024valueincentivized} have proposed offline methods that learn policies against an adversarial reward function, leveraging the DPO simplification~\citep{rafailov2023direct} to avoid full-fledged adversarial training; among these, \citet{cen2024valueincentivized} also introduces an online variant. \citet{eisenstein2024helping} explores reward uncertainty through an ensemble of reward models, showing that ensemble aggregation helps mitigate reward hacking, though it does not fully resolve over-optimization risks.

% --- Preference Learning and Pessimism ---
On the preference-learning front, recent work has relaxed the BTL assumption, either by bypassing the need for a separate preference model~\citep{azar2023general, tang2024generalized} or adopting self-play approaches~\citep{munos2023nash, swamy2024minimaximalist, calandriello2024human, ye2024online}.
\citet{ye2024online} introduces a pessimistic preference objective with unrestricted Nash, and empirically evaluates an uncertainty based exploration method for the preference optimization. 
However, the role of pessimism in the more general preference-learning setting remains largely unexplored. 

\section{Optimal Solution of the Example in Figure~\ref{fig:bad_example}}
\label{sec:bad_example}

Let $\pi_i = \pi(y_i)$ and $p_{23} = p(y_2, y_3)$ and $p_{13} = p(y_1, y_3)$. Then  we can write the objective in this example as 

\begin{align*}
V^\star = \max_{\pi} \min_{\pi'} \min_{p \in\cP} J(\pi, \pi', p)& \\ 
    \textrm{ where } J(\pi, \pi', p) = p(\pi, \pi') = \big( &\,\,
    0.5\pi_1\pi'_1 + 1\cdot\pi_1\pi'_2 + p_{13} \pi_1\pi'_3  \\
    & + 0 \cdot \pi_2 \pi'_1 + 0.5 \pi_2\pi'_2 + p_{23} \pi_2\pi'_3  \\
    &+ (1-p_{13}) \pi_3\pi'_1 
    + (1-p_{23})\pi_3\pi'_2 + 0.5\pi_3\pi'_3 \big).
\end{align*}
\paragraph{Upper-bound on $V^\star$: }
First note that for any $\pi$:
\begin{align*}
    \min_{\pi'} \min_{p \in\cP} J(\pi, \pi', p) \leq 0.5 \pi_3 \qquad \textrm{and} \qquad \min_{\pi'} \min_{p \in\cP} J(\pi, \pi', p) \leq 0.5 \pi_1.
\end{align*}
The first inequality follows by considering the choice $p_{13} = p_{23} = 1$, and the second inequality from considering the choice $\pi'_1 = 1$ and $p_{13} = 1$. Since both bounds hold simultaneously and $\pi_1 + \pi_3 \leq 1$, we can conclude that 
\begin{align*}
    V^\star \leq 0.25.
\end{align*}
%

%Consider any policy $\pi$ such that $\pi_3 = 0$. Then we can simply choose $\pi'_3 = 1$, $P(y_1, y_3) = 0$ and $P(y_2, y_3) = 0$ to obtain an objective value of $0$ for this policy. In fact, we can further conclude that $0.5\pi_3$ is an upper bound on the objective value for any $\pi$ by the same argument more generally. By a similar argument, and choosing $\pi'_1 = 1$, we can also conclude that $0.5\pi_1$ is an upper bound on the objective value for $\pi$. 
\paragraph{Lower-bound on $V^\star$:}
Choosing $\pi_1 = \pi_3 = 0.5$, we see that the objective value can be written as
\begin{align*}
   J(\pi, \pi', p) =& \big( 0.25\pi'_1 + 0.5\pi'_2 + 0.5p_{13} \pi'_3  
   \\&+ 0.5(1 - p_{13}) \pi'_1  + 0.5 (1 - p_{23}) \pi'_2 + 0.25\pi'_3 \big).
\end{align*}
First we observe that the minimum of this quantity is always attained at $p_{23} = 1$ and thus we can ignore the penultimate term.
Consider now two cases:
\begin{itemize}
    \item Case $\pi'_1 \leq \pi'_3$: Then the coefficient of $p_{13}$ is non-negative and the minimum is attained at $p_{13} = 0$. This allows us to simplify the expression further as
\begin{align*}
    \min_{\pi'} \min_{p\in\cP} J(\pi, \pi', p) &= 
    \min_{\pi'} 0.25\pi'_1 + 0.5\pi'_2 + 0.5\pi'_1 + 0.5\pi'_2 + 0.25\pi'_3\\
    &= \min_{\pi'}  0.75\pi'_1 + \pi'_2 + 0.25\pi'_3\\
    &= 0.25
\end{align*}
where we choose $\pi'_3 = 1$ in the last step.
\item Case $\pi'_1 \geq \pi'_3$: Then the coefficient of $p_{13}$ is non-positive and the minimum is attained at $p_{13} = 1$. This gives  
\begin{align*}
\min_{\pi'} \min_{p\in\cP} J(\pi, \pi', p) &=
    \min_{\pi'}  0.25\pi'_1 + 0.5\pi'_2 + 0.5\pi'_3 + 0.5\pi'_2 + 0.25\pi'_3\\
    &= \min_{\pi'} \min_{P\in\cP} 0.25\pi'_1 + \pi'_2 + 0.75\pi'_3\\
    &= 0.25
\end{align*}
where the optimal solution is to choose $\pi'_1 = 1$.
\end{itemize}
Combining both cases, we can conclude that
\begin{align*}
    V^\star \geq 0.25.
\end{align*}

\paragraph{Optimal solution.}
Combining both upper- and lower-bounds, we can conclude that $V^\star = 0.25$ which is attained at $\pi_1 = \pi_3 = 0.5$.

\section{Definition and Analysis of Restricted Nash Policy}
\label{sec:restricted-nash}

\subsection{Restatement and Proof of Lemma~\ref{lem:restricted-nash}} \label{appendix:lem:restricted-nash-proof}
We restate Definition~\ref{def:covered-policy-set} and Lemma~\ref{lem:restricted-nash} over here, but with conversation context ($x$) included in the equations for clarity, hence for this section we redefine $\cP \subseteq \{\X\times\Y\times\Y \to [0,1]\}$. We also use $x\sim\D$ to denote a context $x$ sampled from the offline dataset. 

\begin{definition}[Covered policy set] For a given sampling policy $\pisample \in \Pi$ and constant $C$, the covered policy set $\Pi(\pisample, C)$ with respect to $\pisample$ is the set of policies  such that $\forall \pi,\pi' \in  \Pi(\pisample, C)$ and $\forall p_1,p_2 \in \cP$,
\begin{align}
\begin{split}
    &\E_{x\sim \D, y \sim \pi(\cdot | x), y' \sim\pi'(\cdot | x)} (p_1(x, y, y') - p_2(x, y, y'))^2  \\ &\hspace{0.5cm}\leq C\cdot \E_{x\sim\D,y, y' \stackrel{\mathrm{i.i.d.}}{\sim}\pisample(\cdot | x)} (p_1(x, y, y') - p_2(x, y, y'))^2.
   % \label{eq:support}
\end{split}
\end{align}
% \label{def:covered-policy-set}
\end{definition}

\begin{lemma}[Preference guarantee for the restricted pessimistic Nash policy]
    We denote the restricted pessimistic Nash policy by $\pirespessnash$  from Eq.~\eqref{eq:restricted-pess-nash}, and let $p^\star$ be the ground-truth preference function underlying $\D$. Then we have that for any $\pi\in\Pi(\pisample, C)$ with $C \geq 1$:
    \begin{equation*}
        p^\star(\pirespessnash, \pi) \geq \frac{1}{2} - 2\sqrt{C\epsilon},
    \end{equation*}
    where $\epsilon$ is a bound on how much preference functions in $\cP$ can disagree in total variation under $\pisample$: $\E_{x \sim \D, ( y, y')\stackrel{\mathrm{i.i.d.}}{\sim} \pisample(\cdot | x)} |p_1(x, y, y') - p_2(x, y, y')| \leq \epsilon$, $\forall p_1, p_2 \in \cP$.
% \label{lem:restricted-nash}
\end{lemma}
\begin{proof}[Proof of Lemma~\ref{lem:restricted-nash}]
    Let $\piresnash^\star$ be a restricted Nash solution under the true preference function $p^\star$:
    \begin{align*}
        \piresnash^\star &= \argmax_{\pi\in\Pi(\pisample, C)} \min_{\pi'\in\Pi(\pisample, C)} \EE{x\sim\D, y\sim\pi(\cdot | x), y' \sim \pi'(\cdot | x)}{p^\star(x, y, y')}\\
        &= \argmax_{\pi\in\Pi(\pisample, C)} \min_{\pi'\in\Pi(\pisample, C)} p^\star(\pi, \pi').
    \end{align*}
    We start by noting that $\piresnash^\star$ is solving an anti-symmetric two player zero-sum game and the constraint set $\Pi(\pisample, C)$ is a convex set whenever $\Pi$ is convex. To see this, consider two policies $\pi, \pi' \in \Pi(\pisample, C)$ and $\alpha \in [0, 1]$. Then
    for any $\pi'' \in  \Pi(\pisample, C)$ and $p \in \cP$, we have
    \begin{align*}
         &\E_{x\sim \D, y \sim (\alpha \pi(\cdot | x) + (1-\alpha) \pi'(\cdot | x)), y' \sim \pi''(\cdot | x)} (p_1(x, y, y') - p_2(x, y, y'))^2\\
         &= \alpha \E_{x\sim\D, y \sim \pi(\cdot | x), y' \sim \pi''(\cdot | x)} (p_1(x, y, y') - p_2(x, y, y'))^2
         + (1-\alpha)  \E_{x\sim\D, y \sim \pi'(\cdot | x), y' \sim \pi''(\cdot | x)} (p_1(x, y, y') - p_2(x, y, y'))^2\\
         &\leq C\cdot \E_{x\sim\D, y, y' \stackrel{\mathrm{i.i.d.}}{\sim}\pisample(\cdot | x)} (p_1(x, y, y') - p_2(x, y, y'))^2
    \end{align*}
    where the first inequality uses linearity of expectation and the second that $\pi, \pi' \in \Pi(\pisample, C)$. Thus, $\alpha \pi + (1-\alpha) \pi' \in \Pi(\pisample, C)$ and $\Pi(\pisample, C)$ is convex. 
    As a consequence of this, we have that $\piresnash^\star \in \argmin_{\pi\in\Pi(\pisample, C)} p^\star(\piresnash^\star, \pi)$ and $p^\star(\piresnash^\star, \piresnash^\star) = 0.5$. Let $\pirespessnash' \in \argmin_{\pi\in\Pi(\pisample, C)} \min_{p\in\cP} p(\pirespessnash, \pi)$. Then we have by definition:
    
    \begin{align*}
        p^\star(\pirespessnash, \pi) =& p^\star(\pirespessnash, \pi) - p^\star(\piresnash^\star, \piresnash^\star) + 0.5\\
        \geq& \min_{p\in\cP} p(\pirespessnash, \pirespessnash') - p^\star(\piresnash^\star, \piresnash^\star) + 0.5\\
        \geq& \min_{p\in\cP}\min_{\pi'\in\Pi(\pisample, C)} p(\piresnash^\star, \pi') - p^\star(\piresnash^\star, \piresnash^\star) + 0.5,
    \end{align*}
    where the first inequality is due to the definition of $\pirespessnash'$, and the second follows from the definition of $\hat \pi$. Let $\widetilde\pi \in \argmin_{\pi'\in\Pi(\pisample, C)} \min_{p\in\cP} p(\piresnash^\star, \pi')$. Then we can further write
    \begin{align*}
        p^\star(\pirespessnash, \pi) \geq& \min_{p\in\cP} p(\piresnash^\star, \widetilde\pi) - p^\star(\piresnash^\star, \widetilde\pi) + 0.5\\
        \geq& 0.5 - \min_{p\in\cP}\sqrt{C\E_{x\sim\D, y, y'\iidsim \pisample(\cdot | x)} (p(x, y, y') - p^\star(x, y, y'))^2},
    \end{align*}
    where the first inequality is due to $\piresnash^\star \in \argmin_{\pi'\in\Pi(\pisample, C)} p^\star(\piresnash^\star,\pi')$, and the second inequality follows from Equation~\ref{eq:support}. We can further upper bound this last term using $(p(x, y, y') - p^\star(x, y, y'))^2 = (\sqrt{p(x, y, y')} + \sqrt{p^\star(x, y, y')})^2(\sqrt{p(x, y, y')} - \sqrt{p^\star(x, y, y')})^2 \leq 2^2(\sqrt{p(x, y, y')} - \sqrt{p^\star(x, y, y')})^2$ as
    \begin{align*}
        p^\star(\pirespessnash, \pi)
        \geq& 0.5 - \min_{p\in\cP}\sqrt{4C\E_{x\sim\D, y, y'\iidsim \pisample(\cdot | x)} (\sqrt{p(x, y, y')} - \sqrt{p^\star(x, y, y')})^2}\\
        \geq& 0.5 - 2\min_{p\in\cP}\sqrt{C\E_{x \sim \D} 2\textrm{H}^2(p, p^\star)}\\
        \geq& 0.5 - 2\min_{p\in\cP}\sqrt{C\E_{x \sim \D} 2\textrm{TV}(p, p^\star)}\\
        \geq& 0.5 - 2\sqrt{C\epsilon},
    \end{align*}
    where the first inequality follows from the definition of Hellinger distance, second inequality from the relationship between Hellinger distance and total variation, and the last step is from our definition of $\cP$.
\end{proof}

\subsection{Bounded-likelihood-ratio-based coverage}
\label{app:bounded_likelihood_coverage}
Let $\Pi_{\sqrt{C}} = \{\pi\in\Pi~:~\|\pi/\pisample\|_\infty \leq \sqrt{C}\}$. Then  $\forall \pi, \pi' \in \Pi_{\sqrt{C}}$ and $\forall p_1, p_2 \in \cP$
\begin{align*}
    \mathbb{E}_{x\sim\D, y \sim \pi(\cdot | x), y' \sim \pi'(\cdot | x)}(p_1(x, y, y') - p_2(x, y, y'))^2 &= \mathbb{E}_{x\sim \D, y, y' \iidsim \pisample(\cdot | x)}\left[\frac{\pi(y |x )\pi'(y|x)}{\pisample(y|x)^2}(p_1(x, y, y') - p_2(x, y, y'))^2\right] \\
    &\leq C \cdot \mathbb{E}_{x\sim \D, y, y' \iidsim \pisample(\cdot | x)}(p_1(x, y, y') - p_2(x, y, y'))^2, 
\end{align*}
which implies that $\Pi_{\sqrt{C}} \subseteq \Pi(\pisample, C)$.

\subsection{Covariance-based coverage for linear preferences}
\label{app:covariance_linear}

Suppose $\cP = \{w^T\phi(x,y,y')~:~w\in\cW\}$ be a collection of linear preferences such that $p^\star\in\cP$. Then the coverage condition~\eqref{eq:support} reduces to 
\begin{align*}
    &\E_{y\sim \pi(\cdot | x), y'\sim\pi'(\cdot | x)} ((w_1 - w_2)^\top\phi(x,y,y))^2 \leq C\E_{y, y'\iidsim D_y(\cdot | x)} ((w_1 - w_2)^\top\phi(x,y,y))^2\\
    \Longleftrightarrow & (w_1 - w_2)^\top\Sigma_{\pi, \pi'}(x) (w_1 - w_2) \leq C(w_1 - w_2)^\top\Sigma_{D_y, D_y} (w_1 - w_2),
\end{align*}
where we denote $\Sigma_{\pi, \pi'}(x) =  \E_{y\sim \pi(\cdot | x), y'\sim\pi'(\cdot | x)} \phi(x,y,y)\phi(x,y,y)^\top$. This condition holds whenever we have 
\begin{equation*}
    \sup_{u} \frac{u^\top \Sigma_{\pi, \pi'}u}{u^\top \Sigma_{D_y, D_y} u} \leq C,
\end{equation*}
which is an alignment condition between the covariances that is significantly weaker than the bounded density ratio condition necessitated by the definition of $\Pi_{\sqrt{C}}$.

\section{Proof of Lemma~\ref{lem:exact-obj}}
\label{appendix:nash-restricted-proof}
\begin{proof}
We consider the following objective for $\alpha \in [0, 1]$ and a more general version, where we derive the objective for different values of KL regularization for the main and opponent policy, i.e.,  $\beta, \beta' \in \mathbb{R}^+$ 
\begin{align}
    \max_\pi\min_{p\in\cP} \min_{\pi'} p(\pi,\pi') - \beta \KL{\pi}{\piref} + (1-\alpha)\beta' \KL{\pi'}{ \piref} + \alpha\beta'\KL{\pi'}{\pisample}
\end{align}
Only looking at the inner minimization of $\pi'$, we get
\begin{align}
    &\min_{\pi'} p(\pi,\pi') + (1-\alpha) \beta'  \sum_y \pi'(y) \log \frac{\pi'(y)}{\piref(y)} + \alpha\beta' \sum_y \pi'(y) \log \frac{\pi'(y)}{\pisample(y)}\\
    &\min_{\pi'} p(\pi,\pi') + \beta' \sum_y \pi'(y) \log \frac{\pi'(y)}{\piref(y)^{1-\alpha} \pisample(y)^{\alpha}}
\end{align}
and thus, the optimal solution for $\pi'$ can be written as
\begin{align}
    \pi'_{\star}(y ) = \frac{1}{Z} \piref(y)^{1-\alpha} \pisample(y)^{\alpha} \exp \left( -\frac{1}{\beta'}  p(\pi, y) \right),
\end{align}
with partition function $Z = \sum_y\piref(y)^{1-\alpha} \pisample(y)^{\alpha} \exp \left( -\frac{1}{\beta'}  p(\pi, y) \right)$.
Plugging this back in the objective above gives
\begin{align}
    &\max_\pi\min_{p\in\cP}  p(\pi,\pi'_\star) - \beta \KL{\pi}{\piref} 
    + \beta' \sum_y \pi'_\star(y) \log \frac{\pi'_\star(y)}{\piref(y)^{1-\alpha} \pisample(y)^{\alpha}}\\
    &= \max_\pi\min_{p\in\cP}  p(\pi,\pi'_\star) - \beta \KL{\pi}{\piref} 
    + \beta' \sum_y \pi'_\star(y) \left(-\frac{p(\pi, y)}{\beta'}\right) - \beta' \log Z\\
    \\&=
    \max_\pi\min_{p\in\cP}   - \beta \KL{\pi}{\piref} 
    - \beta' \log Z\\
    &= \max_\pi\min_{p\in\cP}   - \beta \KL{\pi}{\piref} 
    - \beta' \log \sum_y \piref(y)^{1-\alpha} \pisample(y)^{\alpha} \exp \left( -\frac{1}{\beta'}  p(\pi, y) \right)\\
    &=\max_\pi\min_{p\in\cP}   - \beta \KL{\pi}{\piref} 
    - \beta' \log \E_{y \sim \pimixalpha} \exp \left( -\frac{1}{\beta'}  p(\pi, y) \right) + \beta' \log Z',
\end{align}
where $\pimixalpha(y) \propto \piref(y)^{1-\alpha} \pisample(y)^{\alpha}$ and $Z' = \sum_y \piref(y)^{1-\alpha}\pisample(y)^{\alpha}$ is a normalization constant, independent of optimization parameters. Dropping this term gives us an equivalent optimization objective in $\pi$. Setting $\beta' = \beta$ (which is usually the case) completes the proof of the lemma.
\end{proof}

\section{Proof of Lemma~\ref{lem:approx-obj}}
\label{appendix:approx-obj-proof}
\begin{proof}
Consider the log-sum-exp term with $\pimixalpha(y) = \frac{1}{Z'} \piref(y)^{1-\alpha} \pisample(y)^{\alpha}$ and $Z' = \sum_{y} \piref(y)^{1-\alpha} \pisample(y)^{\alpha}$ as
\begin{align*}
    &=\log \E_{y \sim \pimixalpha} \exp \left( -\frac{1}{\beta'}  p(\pi, y) \right)
    \\
    &= \log \E_{y \sim \pi'} \left[ 
    \frac{\pimixalpha(y)}{\pi'(y)}
    \exp \left( -\frac{1}{\beta'}  p(\pi, y) \right)\right]\tag{$\pi'$ arbitrary}
    \\
    &\geq  \E_{y \sim \pi'} \left[ 
    \log\left(\frac{\pimixalpha(y)}{\pi'(y)}
    \exp \left( -\frac{1}{\beta'}  p(\pi, y) \right)\right)\right] \tag{Jensen's inequality}\\
    &= -\frac{1}{\beta'} p(\pi, \pi') + \E_{y \sim \pi'} 
    \log\left(\frac{\pimixalpha(y)}{\pi'(y)}\right) = -\frac{1}{\beta'} p(\pi, \pi') - \KL{\pi'}{\pimixalpha}.
\end{align*}
Setting $\beta' = \beta$ and taking the minimum over $p \in \cP$ yields
\begin{align}
    \min_{p\in\cP} - \log \E_{y\sim\pimixalpha(\piref, \pisample)} \bigg[\exp\big( \frac{-p(\pi, y)}{\beta}\big)\bigg]
    \leq \min_{p\in\cP} \frac{p(\pi, \pi')}{\beta} - \KL{\pi'}{\pimixalpha(\piref, \pisample)}.
\end{align}
Choosing $\pi' = \pimixalpha(\bar \pi, \pisample)$ gives the desired result with $\kappa = -\KL{\pimixalpha(\bar \pi, \pisample)}{\pimixalpha(\piref, \pisample)}$.

\end{proof}

\section{\prpolong}\label{appendix:prpo}
A straightforward way to simplify our general preference-based algorithm \alg\ is to replace the general preference function with a BTL reparameterization, $\pbt(r)$, that uses an underlying reward function $r$. This substitution yields the objective in \eqref{eq:prpo}. We provide modified pseudo-code in Algorithm~\ref{alg:prpo}, where we also use the preference learning rate $\eta_p$ as the reward function’s learning rate.

Building on \citet{azar2023general}, we can further consider a monotonically increasing function $\Psi: [0,1] \rightarrow \mathbb{R}$, leading to the modified objective:
\begin{align*}
    \max_\pi \min_r 
    \EE{y\sim \pi,\; y'\sim \pimixalpha(\bar\pi,\;\pisample)}{
      \Psi\bigl(\pbt(y,\;y';\;r)\bigr)
    }
    \;-\;\beta \KL{\pi}{\piref}
    \;-\;\lambda\,\mathrm{KL}_{\pisample}\bigl(\pbt(\rmle)\;\|\;\pbt(r)\bigr).
\end{align*}
When $\Psi$ is the identity function, we recover \prpo. By contrast, setting 
$\Psi(q)\;=\;\ln\!\bigl(q/(1-q)\bigr)$ produces the objective
\begin{align*}
    \max_\pi \min_r 
    \EE{y\sim \pi,\; y'\sim \pimixalpha(\bar\pi,\;\pisample)}{
      r(y)\;-\;r(y')
    }
    \;-\;\beta \KL{\pi}{\piref}
    \;-\;\lambda\,\mathrm{KL}_{\pisample}\bigl(\pbt(\rmle)\;\|\;\pbt(r)\bigr).
\end{align*}
This latter form matches existing pessimistic reward-based methods \citep{fisch2024robust, liu2024provably}, although those works often fix the opponent (rather than using $\pimixalpha$) and employ a log-likelihood term to maintain a version space of plausible reward functions. Both \citet{liu2024provably} and \citet{fisch2024robust} circumvent the inner minimization by solving it in closed form.

\begin{algorithm}[tb]
   \caption{\prpolong (\prpo($\alpha$))}
   \label{alg:prpo}
\begin{algorithmic}
   \STATE {\bfseries Hyperparameters:} Mixing coefficient $\alpha$, policy regularization coefficient $\beta$, preference regularization coefficient $\lambda$, exponential moving average parameter $\gamma$, learning rates $(\eta_p, \eta_\pi)$
   \STATE {\bfseries Initialize:} $\bar{\pi}_1 = \pi_1 = \piref$, $r_1 = \rmle$
   \FOR{$t=1, 2, \dots$}
      \STATE Set $\pimixalpha \propto \bar{\pi}^{1-\alpha}_t \pisample^{\alpha}$ as mix of $\piref$ and EMA $\bar{\pi}_t$ for restricted Nash, 
      \STATE Approximate current objective \eqref{eq:prpo}:
      \begin{align*}
          \Jprpo{\alpha}(\pi_t, r_t) \doteq 
          \pbt\Big(\pi_t, \pimixalpha;r_t\Big)  - \beta \KL{\pi_t}{\piref}  - \lambda \mathrm{KL}_{\pisample}\left(\pbt(\rmle)\| \pbt(r)\right)
        %   \EE{y,y'\sim \piref}{\KL{\pmle(y,y')}{p(y,y')}}
      \end{align*}
      \STATE Update $\pi_{t+1} \leftarrow \pi_t + \eta_\pi \frac{\partial \Jprpo{\alpha}(\pi, r_t)}{\partial \pi} \bigg|_{\pi = \pi_t}$
      \STATE Update $r_{t+1} \leftarrow r_t - \eta_p \frac{\partial \Jprpo{\alpha}(\pi_t, r)}{\partial r} \bigg|_{r = r_t}$
      \STATE Update $\bar{\pi}_{t+1} \leftarrow \gamma \pi_t + (1-\gamma) \bar{\pi}_t$
   \ENDFOR
\end{algorithmic}
\end{algorithm}

\section{Tabular Experiments}\label{appendix:tabular-experiments}
% \paragraph{Tabular experiments.}
To illustrate our approach and evaluate the proposed objectives, we conduct experiments in a tabular setting with three possible outputs $y \in \{y_1, y_2, y_3\}$,  and a ground-truth preference matrix \(p^\star\). We vary the probability of sampling $y_3$ from $0.0\rightarrow 0.2$, distributing the remaining probability equally among $(y_1, y_2)$. Thus, $y_3$ is consistently the under-sampled output.
Under each sampling policy, we collect 500 action pairs and use $p^\star$ to sample their pairwise preferences, forming a preference dataset. From this dataset, we estimate the empirical preference model $\pmle$. We then define an uncertainty set $\cP(\pmle, c)\textsuperscript{\footnotemark}\subset \{\Y\times\Y\ \to [0,1]\} $\footnotetext{Note that this differs from our earlier definition in~\eqref{eq:uncertainity-set}, which took the preference dataset $\D$ as the first argument.} around $\pmle$ by enumerating all preferences satisfying:
\begin{align*}
    \cP(\pmle, c) = \Bigl\{ p  : \pmle^{c-}(y,y') \leq p(y,y') \leq \pmle^{c+}(y,y') \Bigr\},
\end{align*}
$\forall y\neq y'$ where,
\begin{align*}
    &\pmle^{c-}(y,y') \coloneqq \max\!\bigl(\pmle(y,y') - c\,\sigma(y,y'),\,0\bigr), \\
    &\pmle^{c+}(y,y') \coloneqq \min\!\bigl(\pmle(y,y') + c\,\sigma(y,y'),\,1\bigr),
\end{align*}
and $\sigma(y,y')$ is the empirical standard deviation. Note that, $\cP(\pmle, 0)$ is a singleton set $\{\pmle\}$.
% based on the samples.
We then optimize the objective in \eqref{eq:restricted-objective-logsumexp} (for $\alpha=0.1$ and $\alpha=0$) via a brute-force search over the main policy, the opponent policy, and the version space of preference models. We specify $\piref$ to the uniform random policy. In each case, we choose the smallest $c^\star$ such that $p^\star$ lies in $\cP(\pmle, c^\star)$:
\begin{align*}
c^\star = \argmin_{c} \{ c \mid p^\star \in \cP(\pmle, c) \}.
\end{align*}

\textbf{Brute-force optimization:} 
We perform a grid search over the main policy, the opponent policy, and all possible preference matrices in the version space. Each policy is discretized into 11 points per action, resulting in $11^3 = 1331$ possible policies for each player. Similarly, we discretize each entry of the preference matrix between $\pmle^{c-}$ and $\pmle^{c+}$ into 11 points. Because the matrix is fully specified by three parameters, this again yields $11^3 = 1331$ possible matrices to search over. To calculate the minimum preference over preferred actions, we drop any action with a probability below $0.05$ for the restricted action set.\looseness=-1

\begin{figure}
    \centering
    \includegraphics[width=0.32\textwidth]{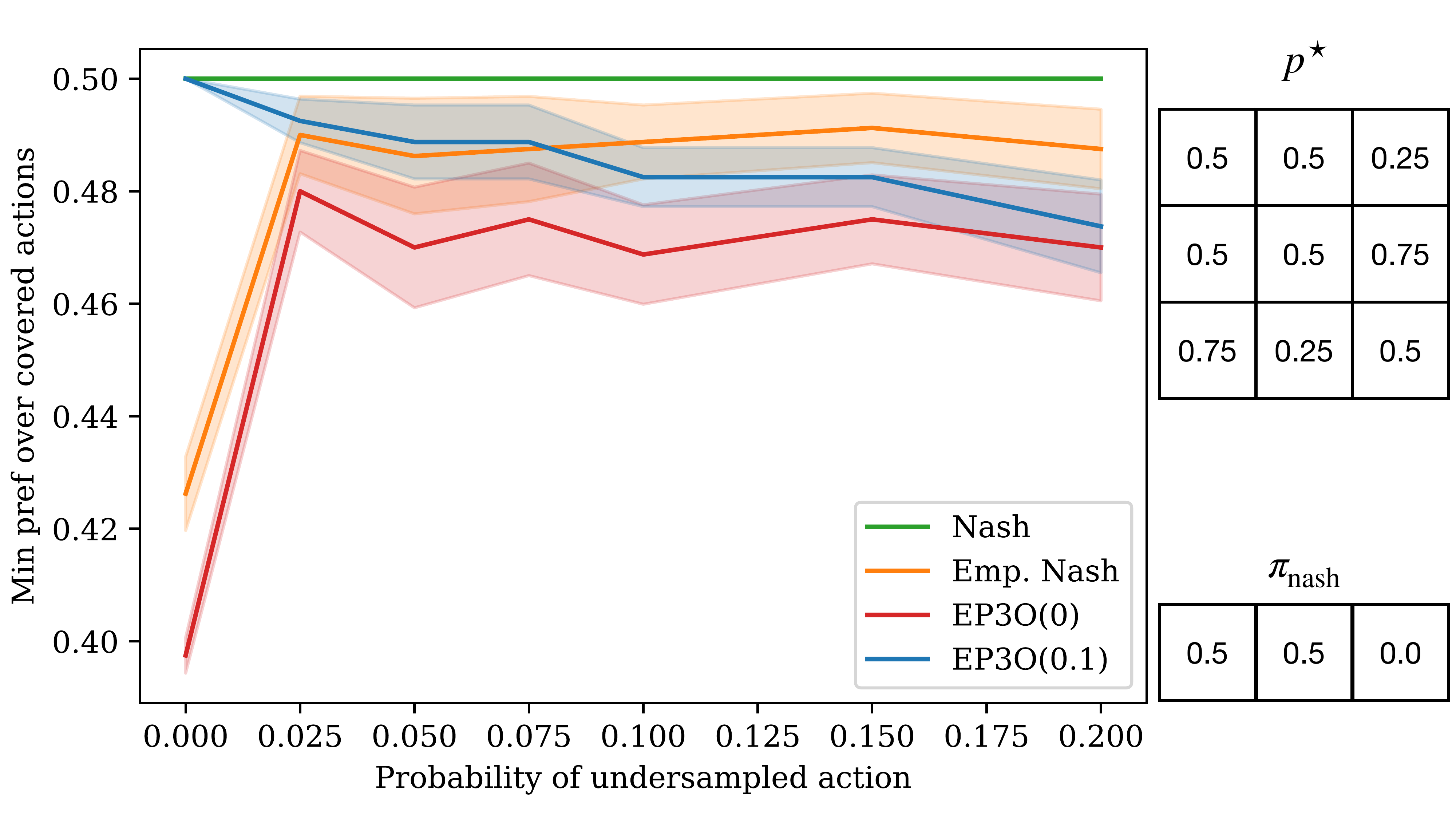}
    \includegraphics[width=0.32\textwidth]{figures/tabular_nash_0_1.pdf}
    \includegraphics[width=0.32\textwidth]{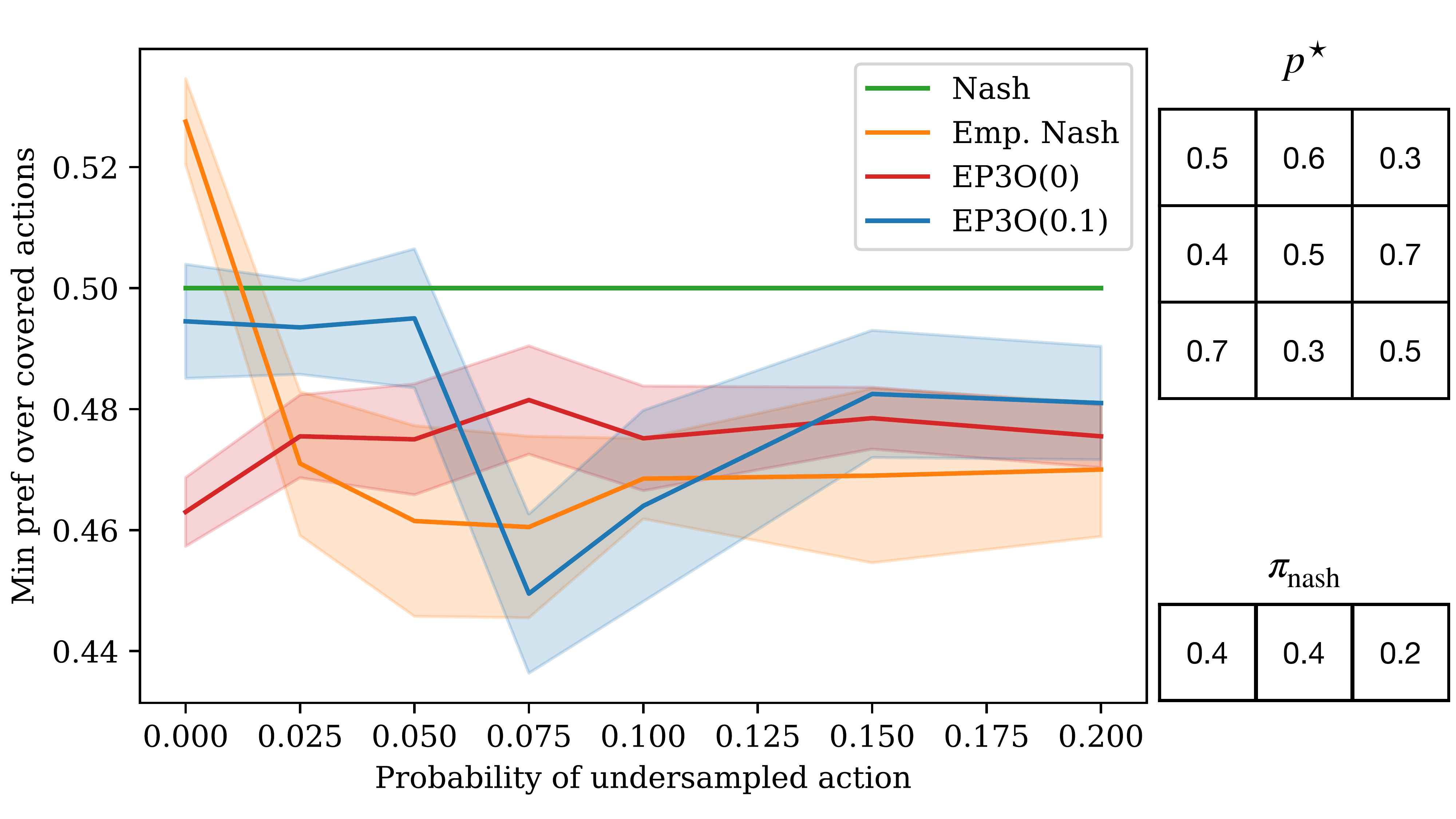}
    \includegraphics[width=0.32\textwidth]{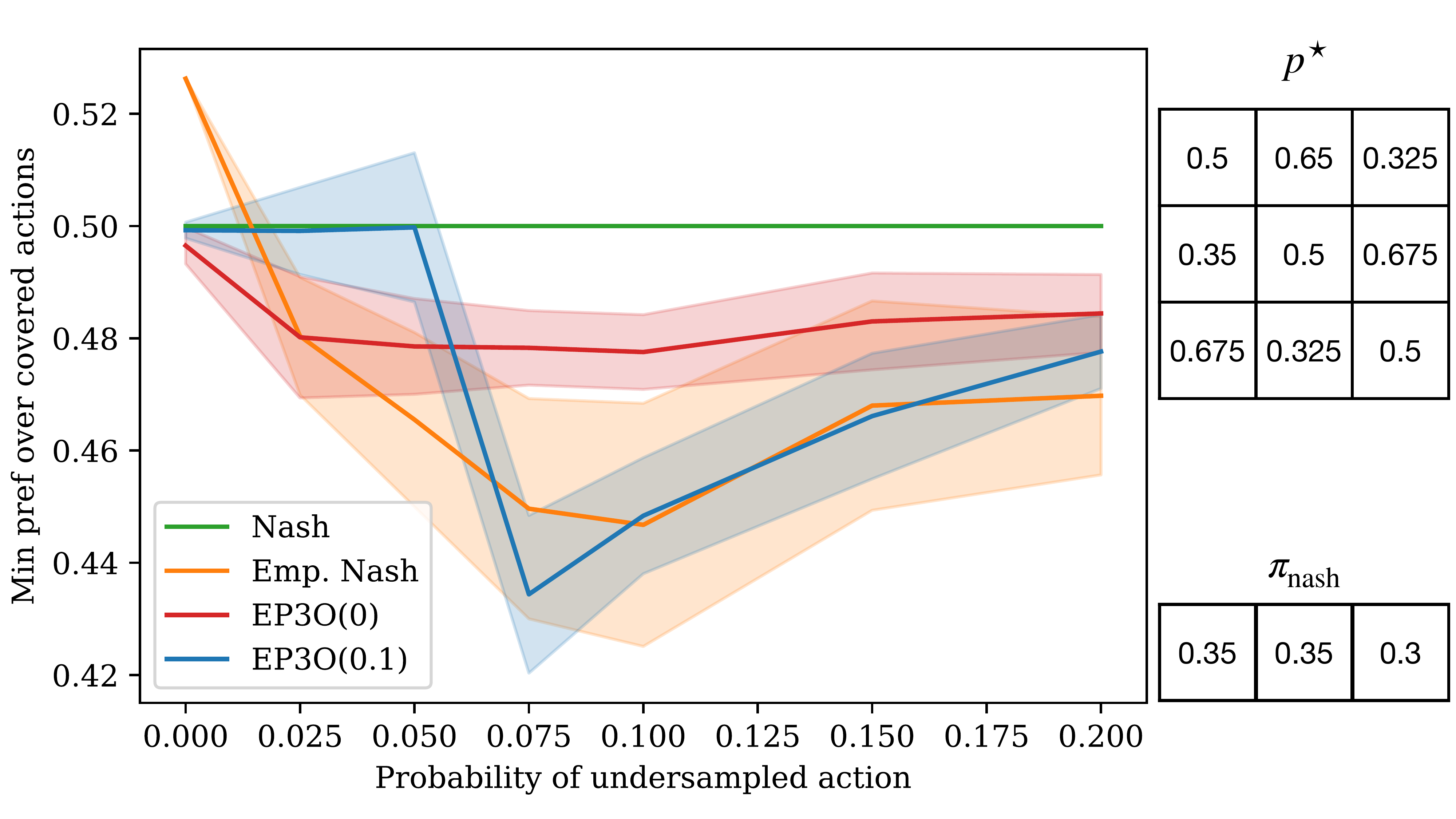}
    \includegraphics[width=0.32\textwidth]{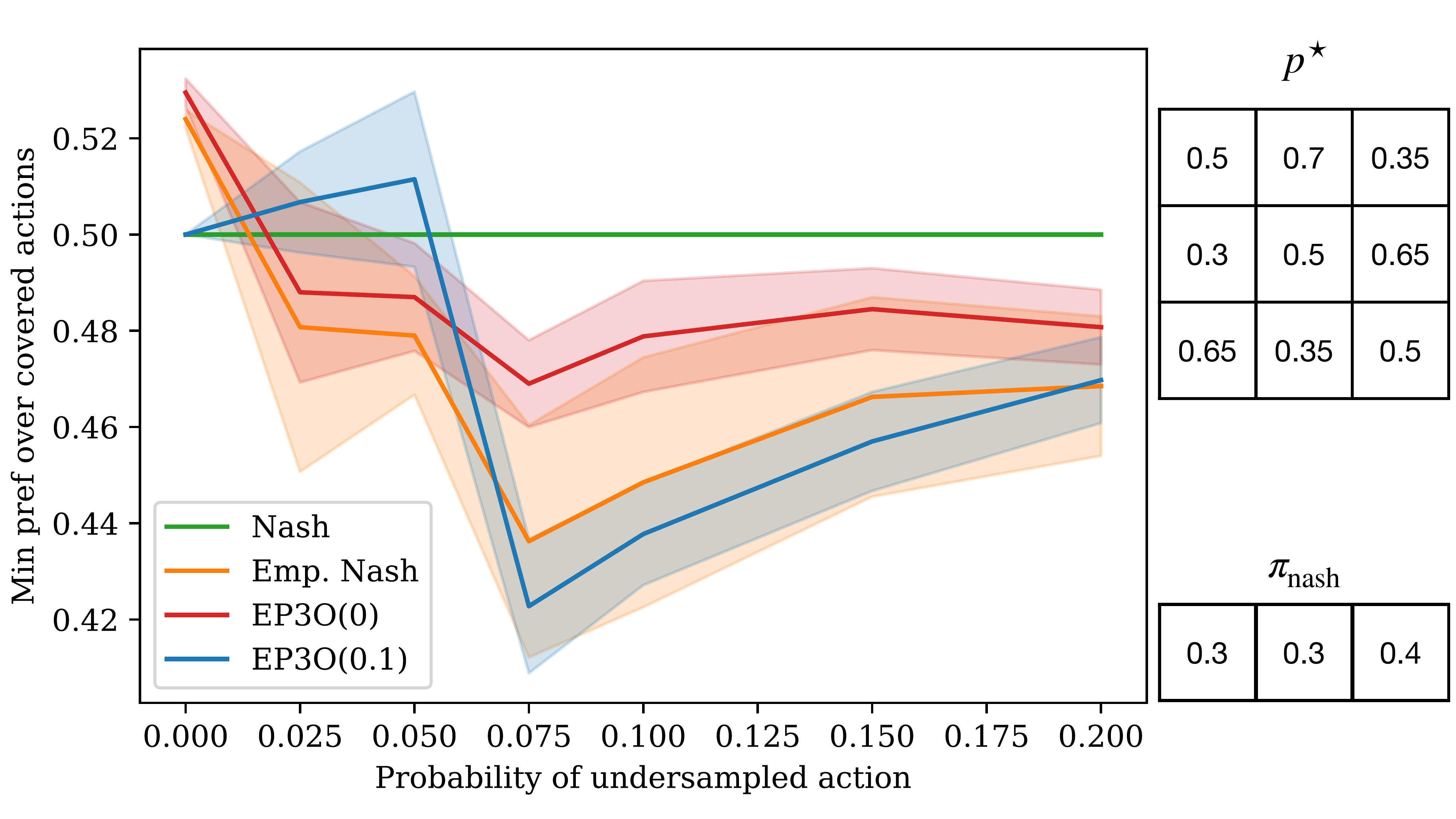}
    \includegraphics[width=0.32\textwidth]{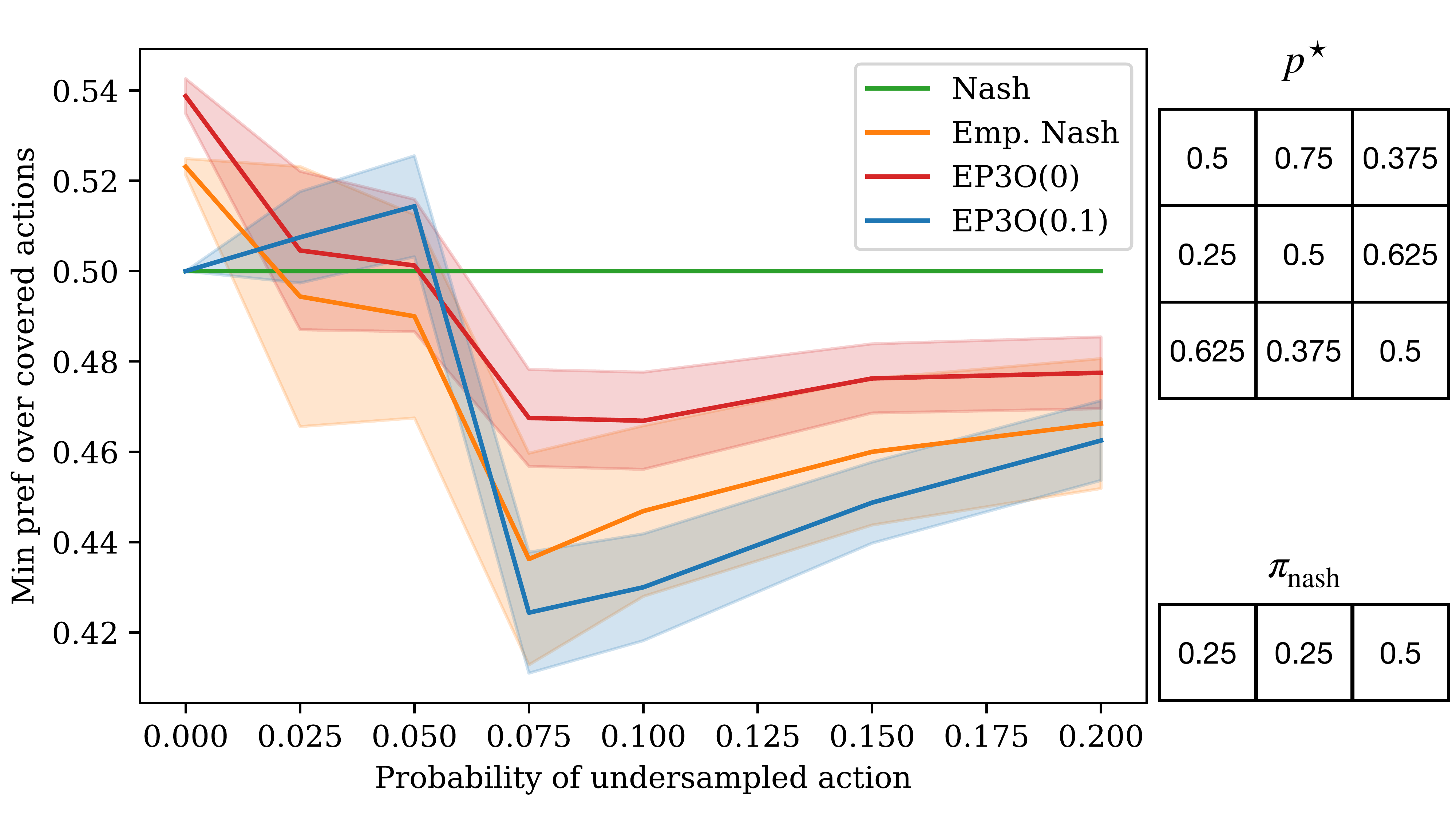}
    \includegraphics[width=0.32\textwidth]{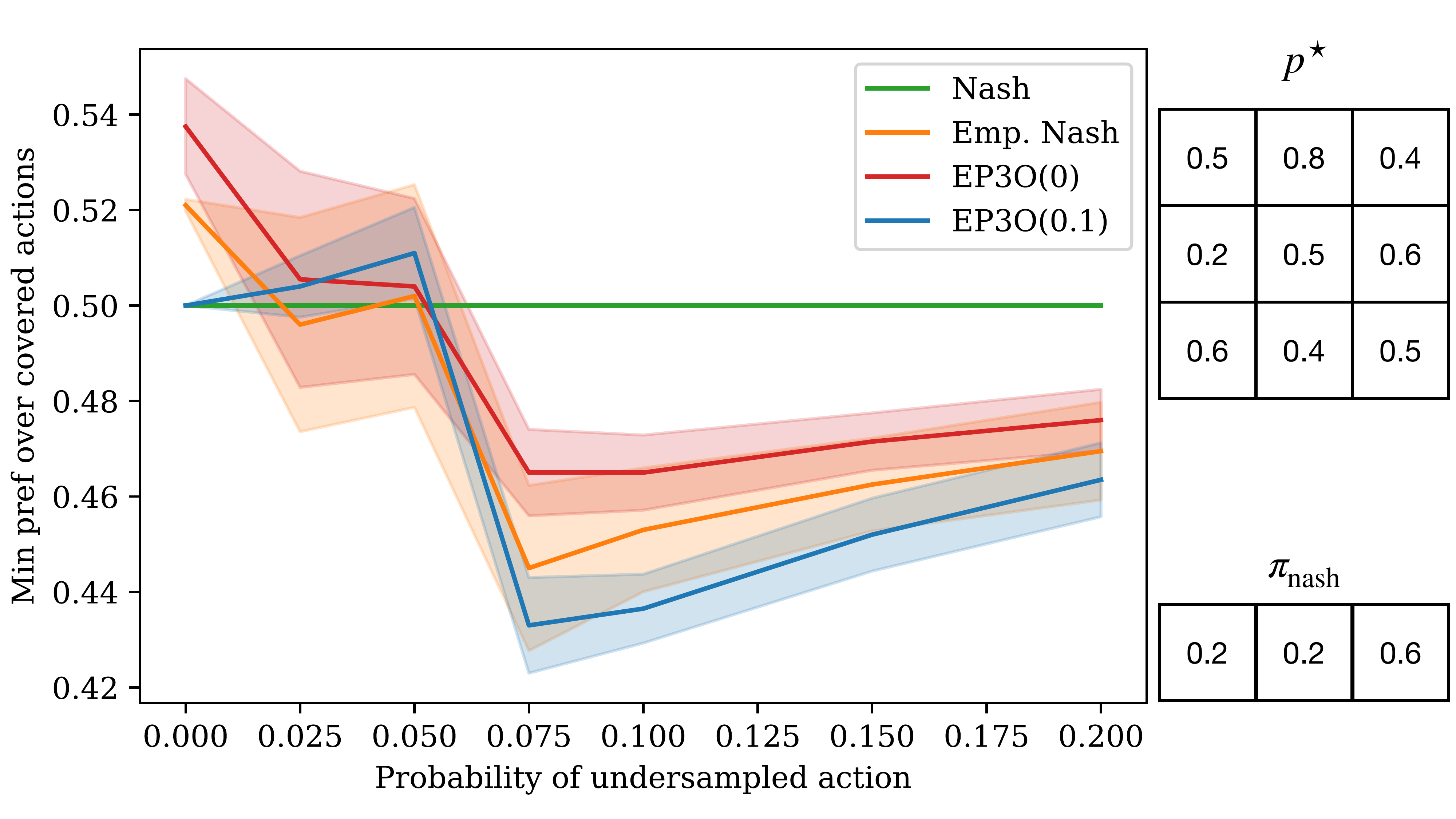}
    \includegraphics[width=0.32\textwidth]{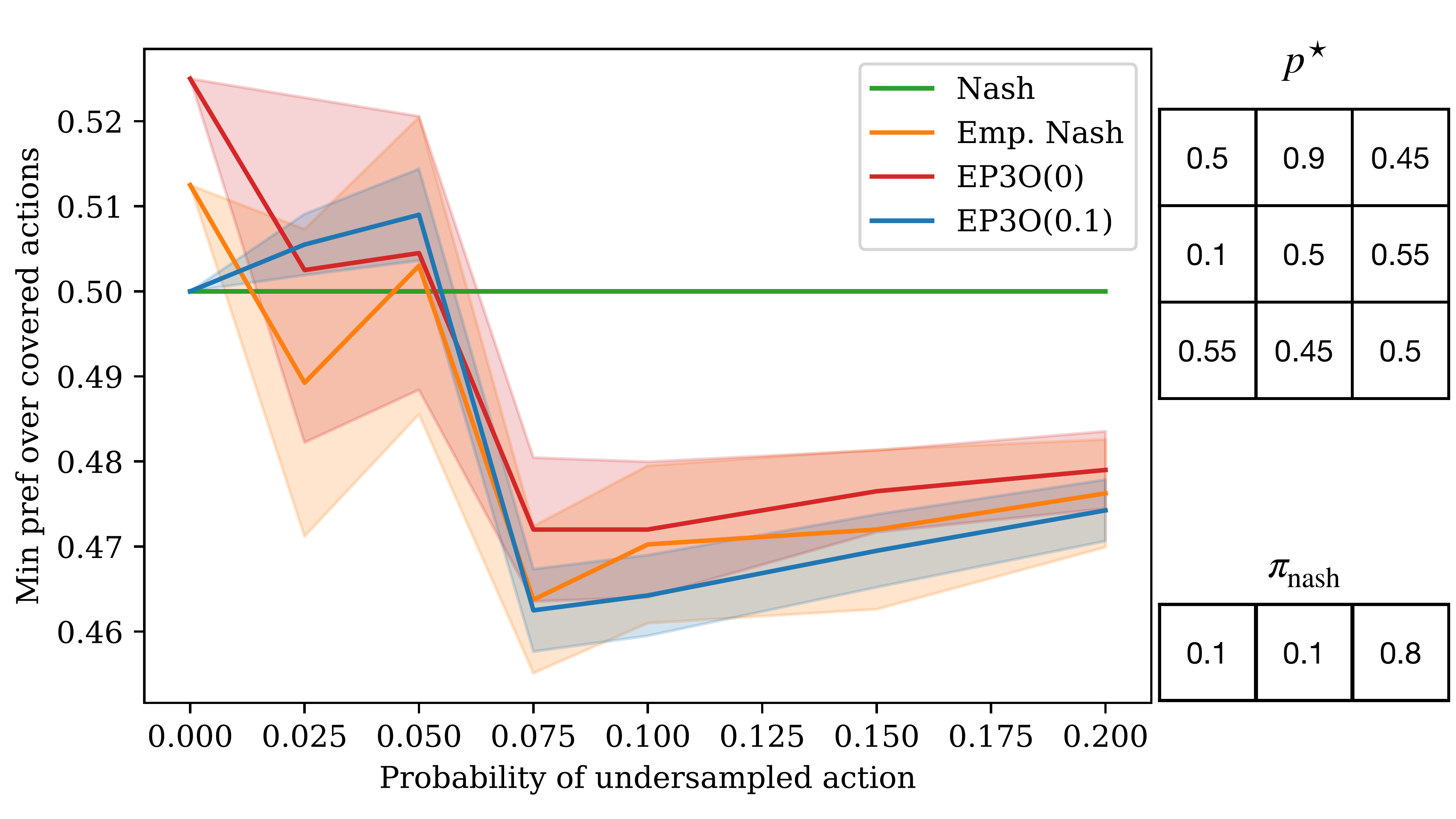}
    \includegraphics[width=0.32\textwidth]{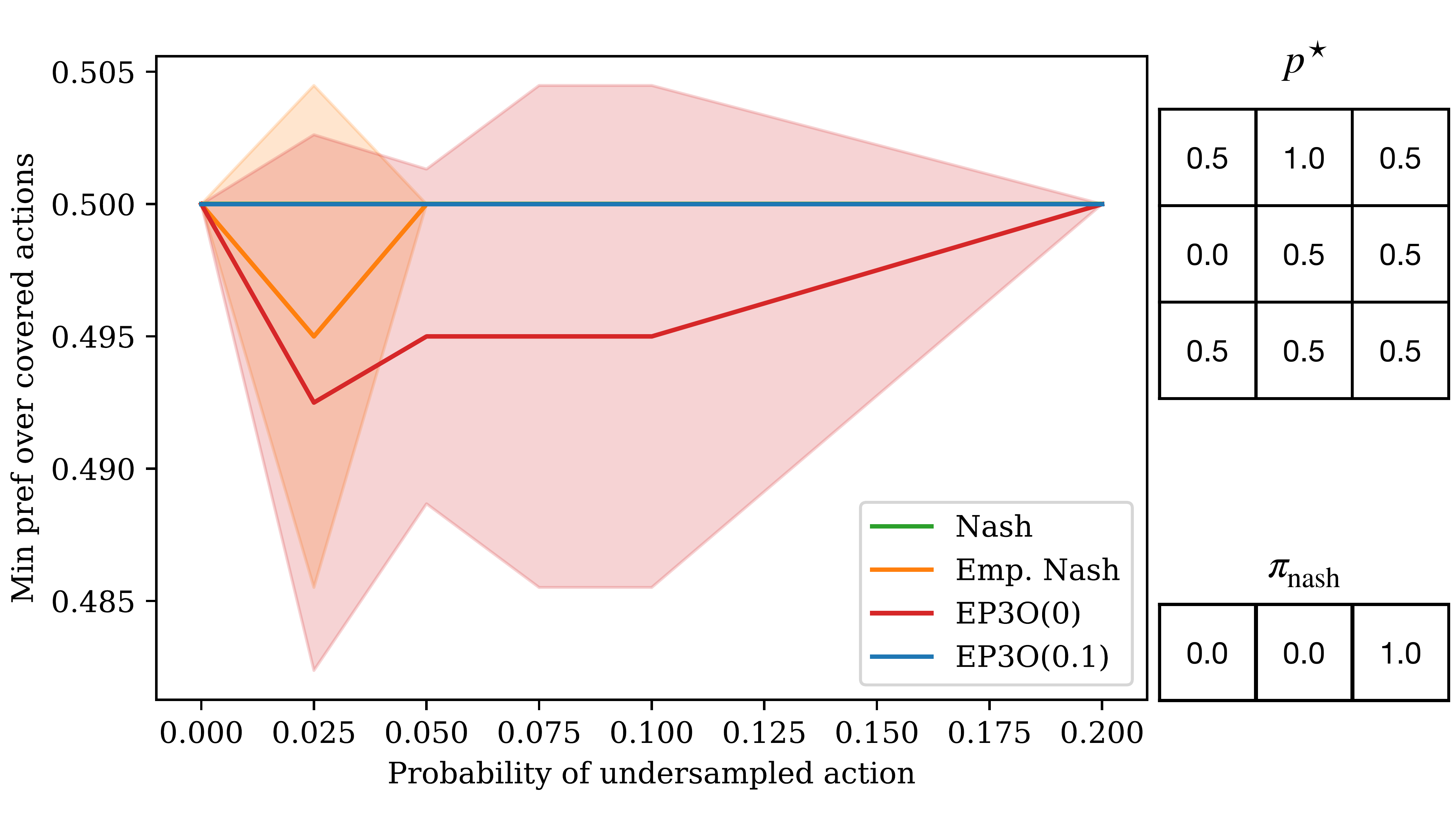}
    \caption{\textbf{Tabular experiments} (Continued from Figure~\ref{fig:tabular-experiments})
    Comparison of the different objective functions with an explicit search over the main policy, opponent policy, and version space.
    The $X$-axis shows the probability assigned to the under-sampled output ($y_3$), from $0.0 \rightarrow 0.2$.
    The $Y$-axis indicates the minimum preference of the policy found over all covered actions (higher is better). Each plot corresponds to a differernt $p^\star$ setting.
    Each figure also shows the ground-truth preference matrix $p^\star$ (on the top-right) and the corresponding $\pinash$ (bottom-right) to each plot. Results are averaged over 10 random seeds, shaded areas represent $\pm 2 \times$ standard error. EP3O(0.1) corresponding to the restricted Nash formulation consistently does well, particularly when the undersampled output is dispreferred and its sampling rate is very low (left part of plot).
    }
    \label{appendix:fig:tabular-experiments}
\end{figure}

\section{Experiments}\label{app:experiments}
\begin{figure*}
    \centering
    \includegraphics[height=1.5in]{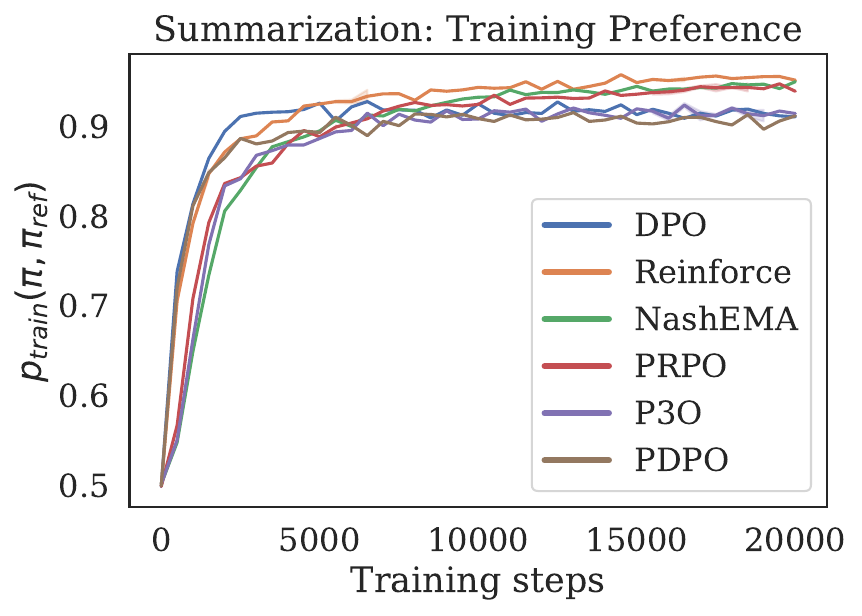}
    \includegraphics[height=1.5in]{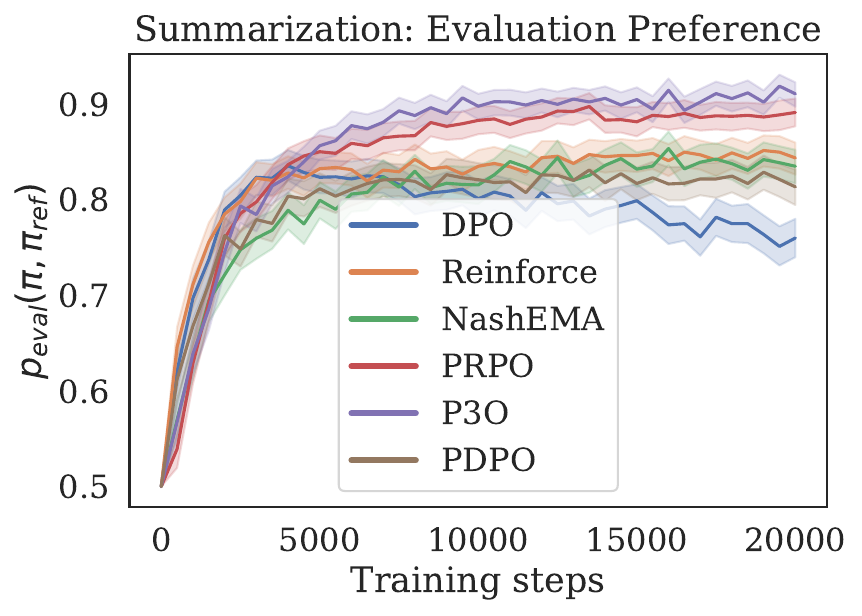}
    \caption{
    \textbf{Training vs.\ evaluation dynamics for summarization.} 
    The horizontal axis shows the number of training steps, and the vertical axis shows the preference of the learned policies against $\piref$. We use the same hyperparameters for each method as in Figure~\ref{fig:eval_pref_gemini}.
    (\textbf{Left}) The training preference is measured by $\pmle$ for preference-based methods and by $\pbt(\rmle)$ for RLHF methods.
    (\textbf{Right}) The evaluation preference measured is by Gemini~1.5~Flash (repeated from Figure~\ref{fig:eval_pref_gemini}).\looseness=-1
    }
    \label{fig:learning-dynamics-summarization}
\end{figure*}

\begin{figure*}
    \centering
    \includegraphics[height=1.5in]{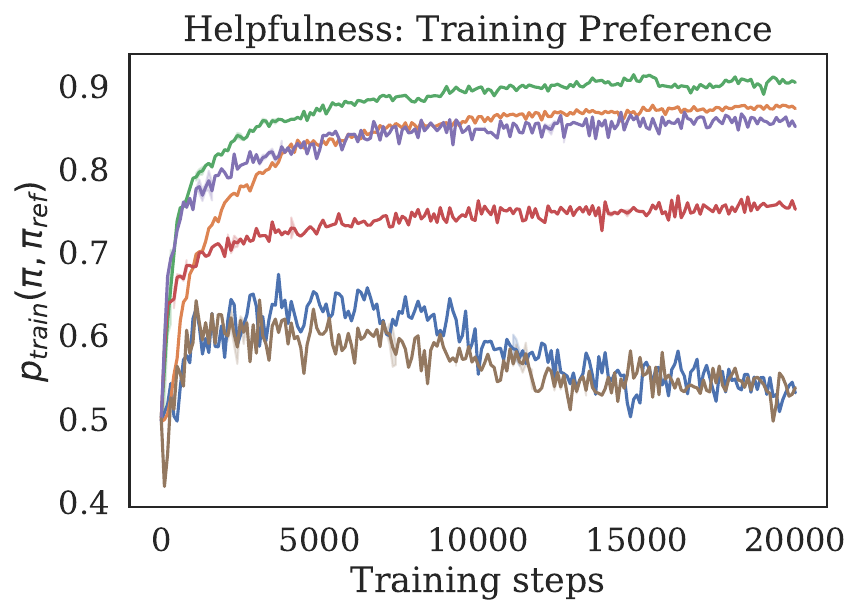}
    \includegraphics[height=1.5in]{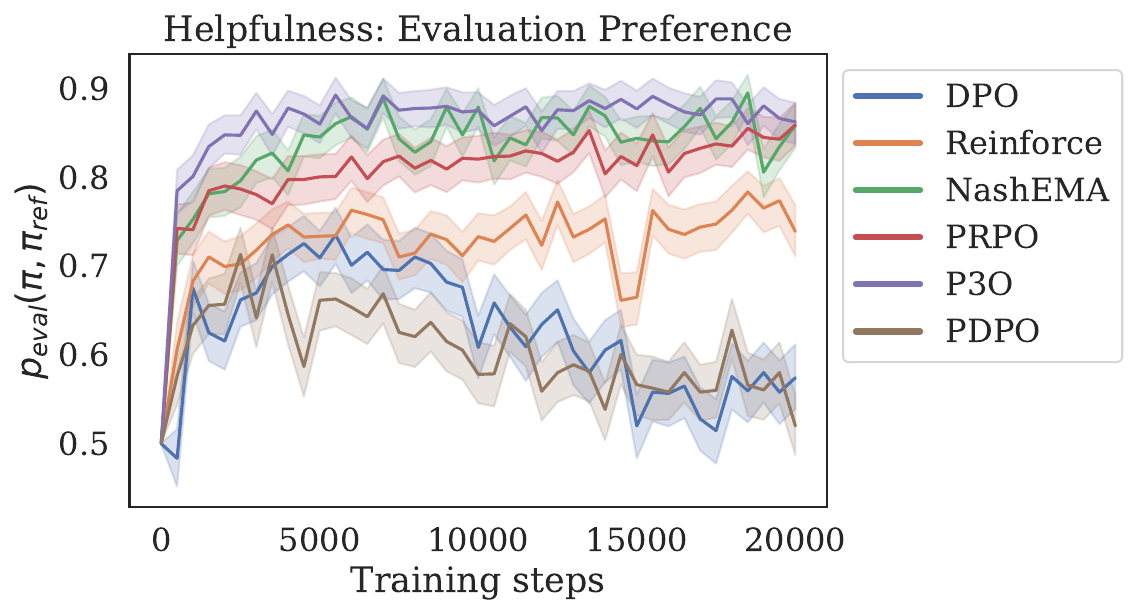}
    \caption{
    \textbf{Training vs.\ evaluation dynamics for helpfulness.} 
    The horizontal axis shows the number of training steps, and the vertical axis shows the preference of the learned policies against $\piref$. We use the same hyperparameters for each method as in Figure~\ref{fig:eval_pref_gemini}.
    (\textbf{Left}) The training preference is measured by $\pmle$ for preference-based methods and by $\pbt(\rmle)$ for RLHF methods.
    (\textbf{Right}) The evaluation preference is measured by Gemini~1.5~Flash.
    }
    \label{fig:learning-dynamics-helpfulness}
\end{figure*}
Figure~\ref{fig:learning-dynamics-summarization} and Figure~\ref{fig:learning-dynamics-helpfulness} presents the learning curves for different methods over a period of either $20,\!000$ training steps or $10,\!000$ training steps (for the larger 8B models). As we note in the curves,  the far left side corresponds to the starting point where $\pi_1 \doteq\piref$, and hence the initial preference is 0.5. In the left figure we can see that all methods consistently seem to be improving on the training reward, where in fact REINFORCE actually seems to be doing \emph{better} than pessimistic methods. However,  that ordering is not followed when evaluated with the much bigger eval model (i.e., Gemini 1.5), as seen in the right figure, wherein pessimistic methods (\alg, \prpo) outperform the standard RLHF methods, and do not degrade over time.\looseness=-1
\begin{figure*}
    \centering
    \includegraphics[height=1.5in]{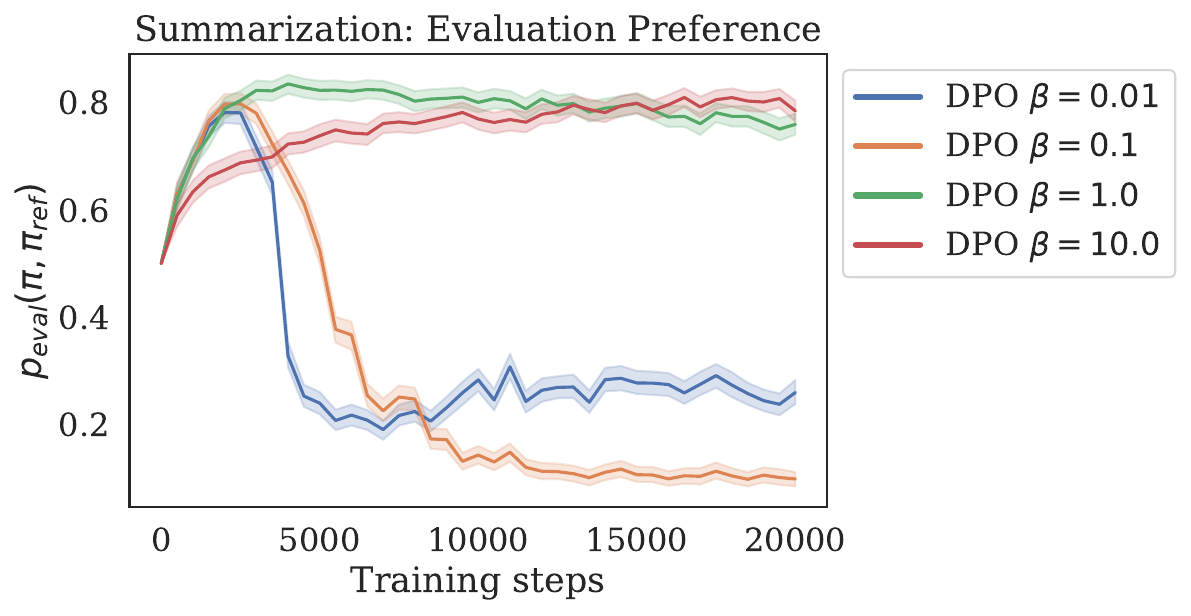}
    \includegraphics[height=1.5in]{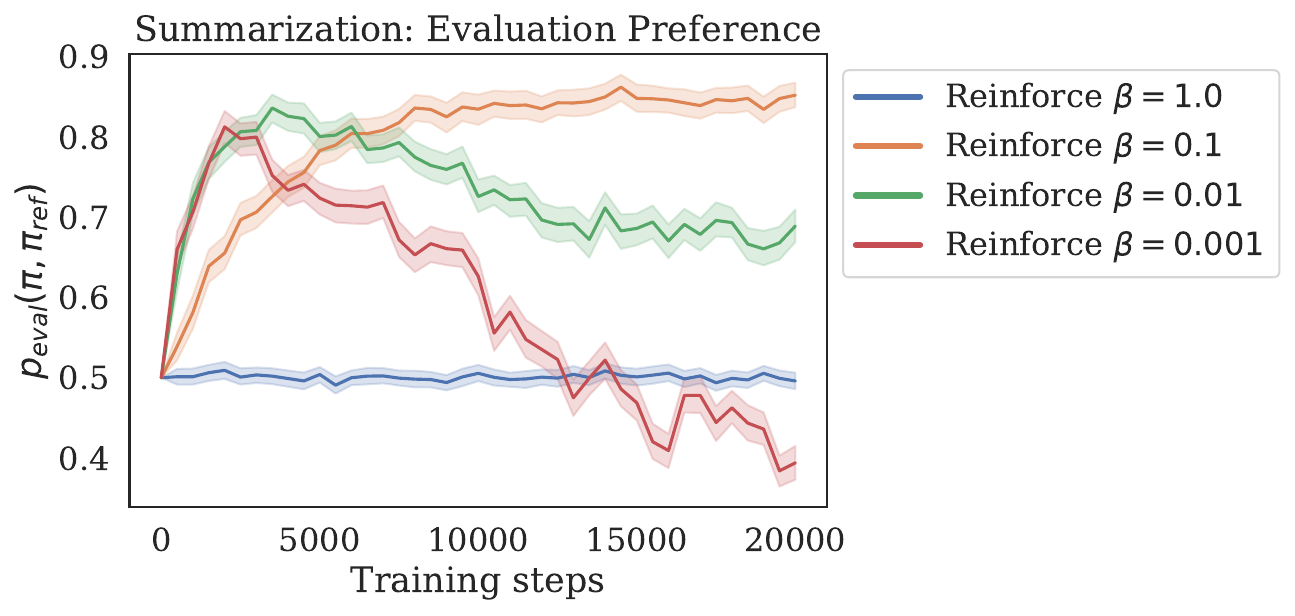}
    \includegraphics[height=1.5in]{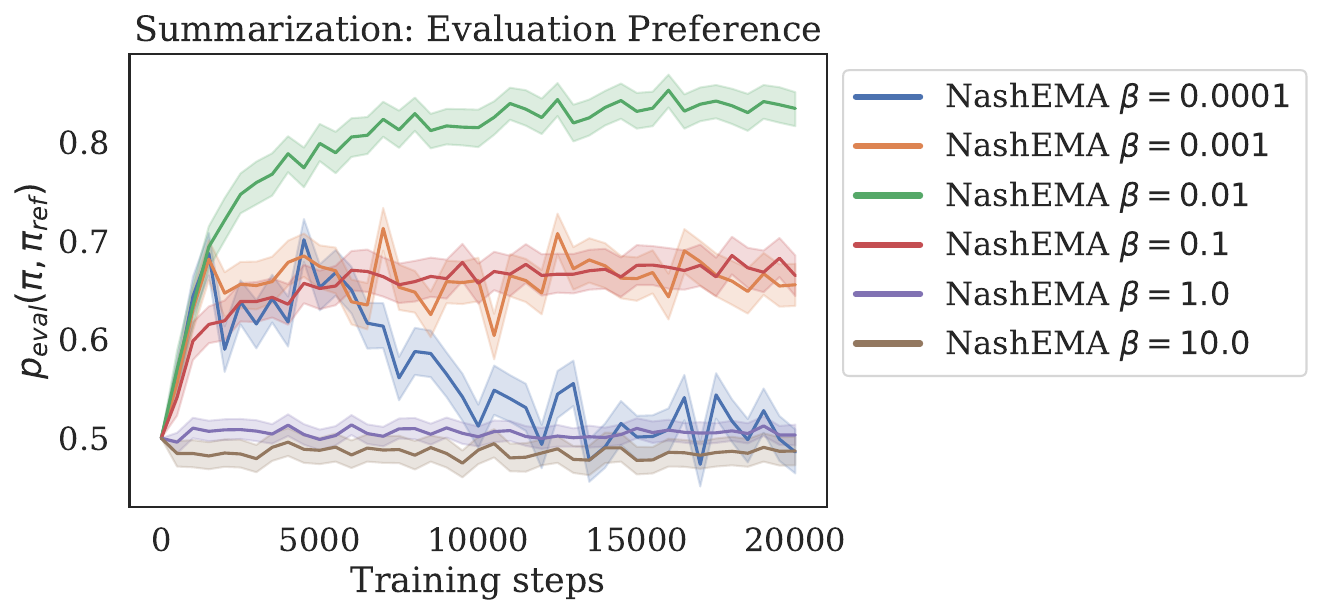}
    \caption{
    \textbf{Ablation of RLHF methods for summarization.}
    The horizontal axis indicates the number of training steps, and the vertical axis shows the preference of the learned policies against $\piref$, as measured by Gemini~1.5~Flash. Hyperparameters are selected for the best evaluation performance.
    (\textbf{Left}) Effect of $\beta$ (KL regularization) on \dpo.
    (\textbf{Right}) Effect of $\beta$ (KL regularization) on REINFORCE.  (\textbf{Bottom}) Effect of $\beta$ (KL regularization) on Nash-EMA.
    Shaded areas represent $95\%$ confidence bands of the evaluation.
    }
    \label{fig:rlhf-ablation-summarization}
\end{figure*}

\begin{figure*}
    \centering
    \includegraphics[height=1.5in]{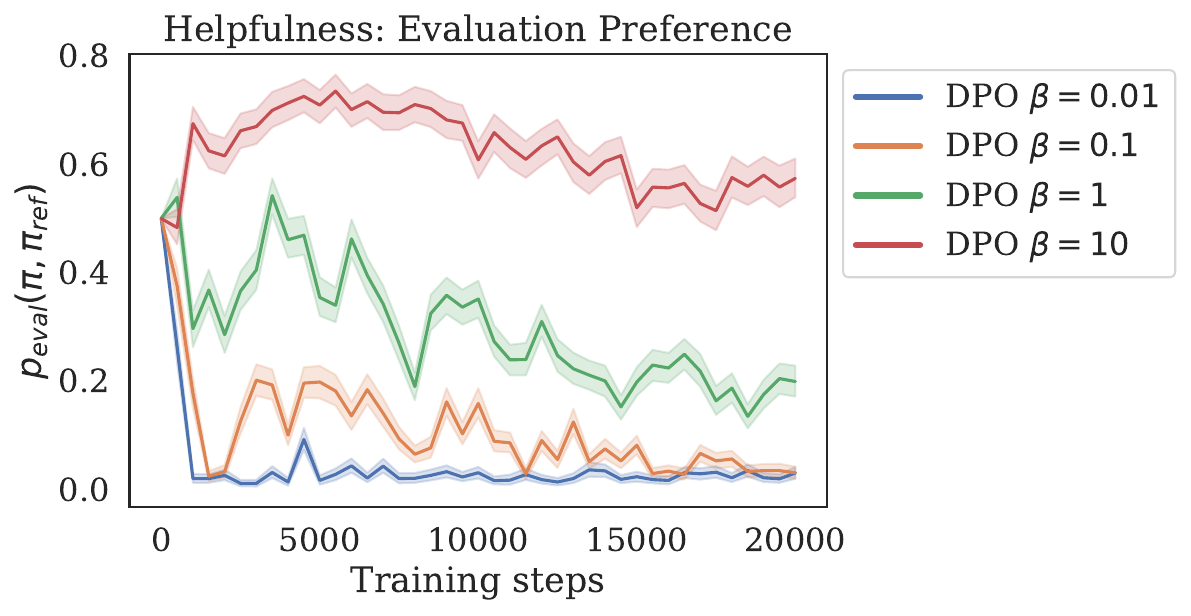}
    \includegraphics[height=1.5in]{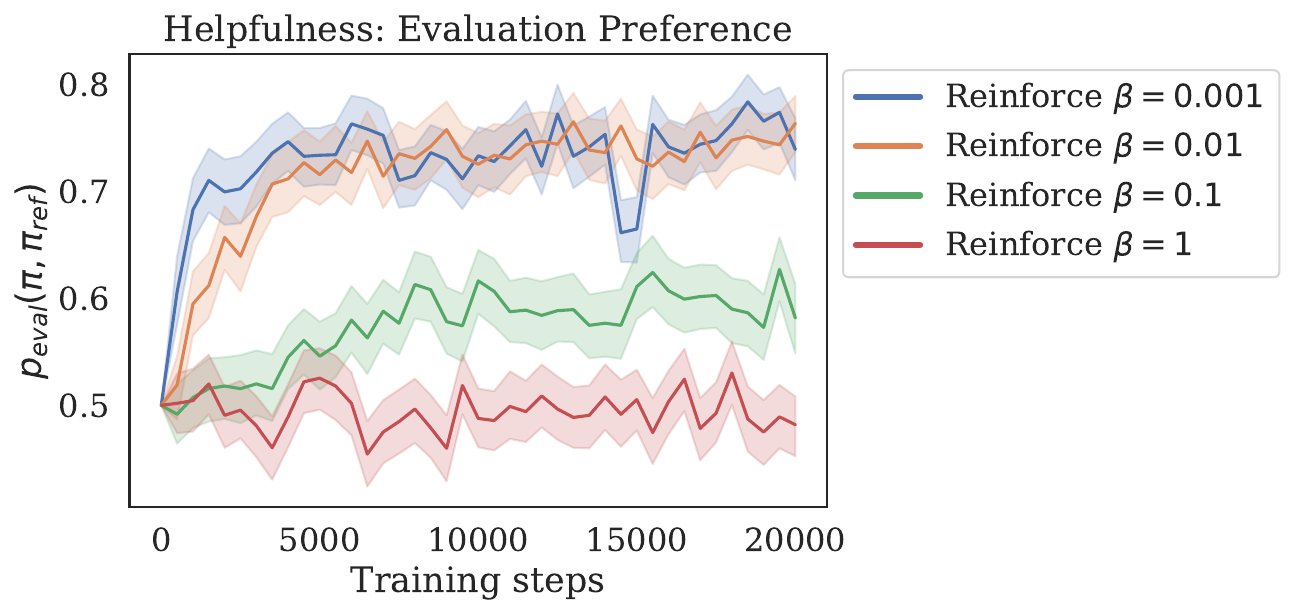}
    \includegraphics[height=1.5in]{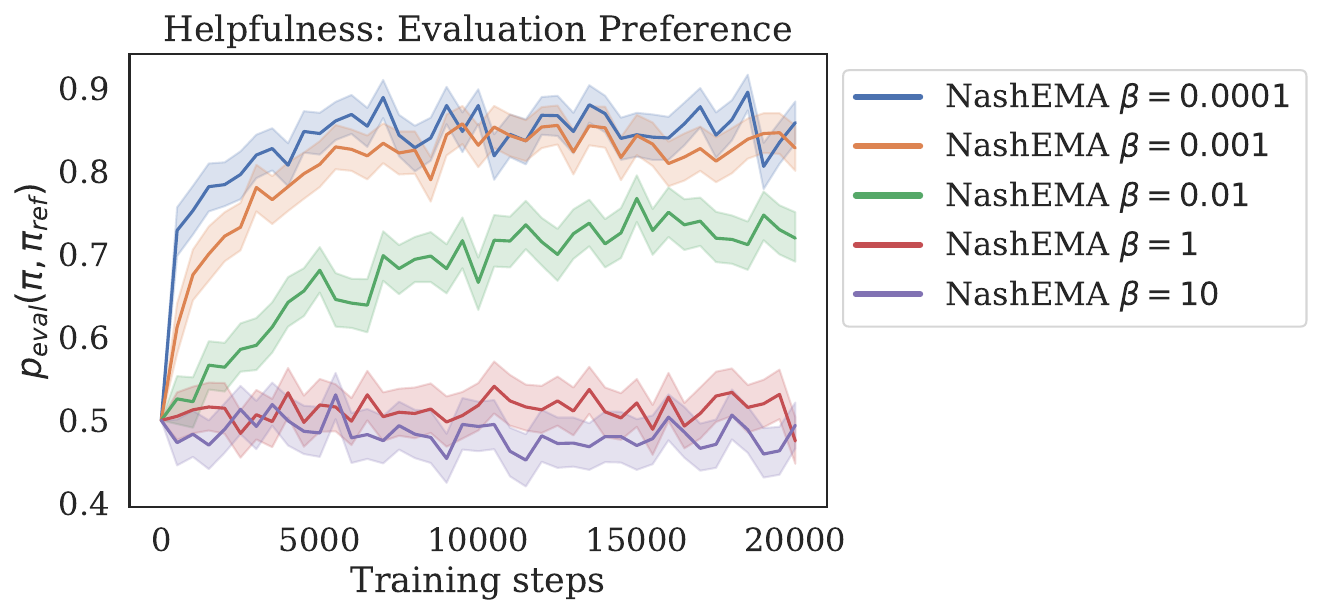}
    \caption{
    \textbf{Ablation of RLHF methods for helpfulness.}
    The horizontal axis indicates the number of training steps, and the vertical axis shows the preference of the learned policies against $\piref$, as measured by Gemini~1.5~Flash.
    (\textbf{Left}) Effect of $\beta$ (KL regularization) on \dpo.
    (\textbf{Right}) Effect of $\beta$ (KL regularization) on REINFORCE. (\textbf{Bottom}) Effect of $\beta$ (KL regularization) on Nash-EMA.
    Shaded areas represent $95\%$ confidence bands of the evaluation.
    }
    \label{fig:rlhf-ablation-helpfulness}
\end{figure*}

% Reward hacking figure for reinforce
\begin{figure*}
    \centering
    \includegraphics[width=0.45\textwidth]{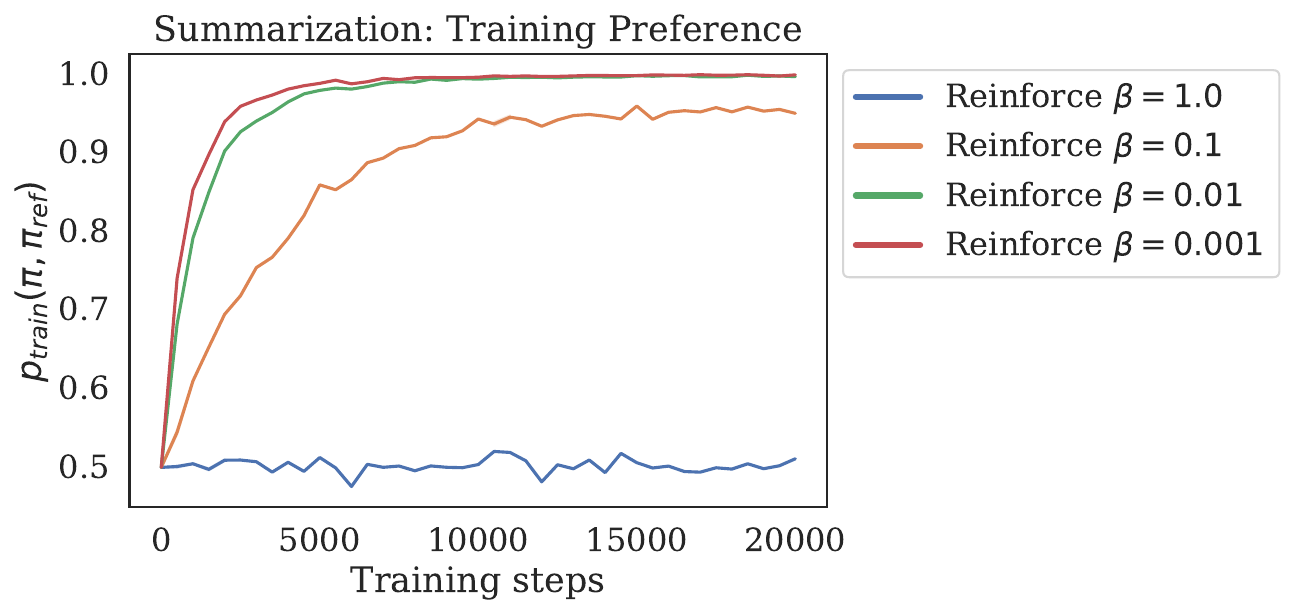}
    \includegraphics[width=0.45\textwidth]{figures/tldr_reinforce_sweep_peval_training_curve.pdf}
    \caption{
    \textbf{Reward hacking in the case of REINFORCE on summarization.}
     The horizontal axis shows the number of training steps, and the vertical axis shows the preference of the learned policies against $\piref$ using $\pbt(\rmle)$ (\textbf{left}) and Gemini~1.5~Flash (\textbf{right}). 
     REINFORCE exhibits reward hacking particularly for $\beta \in \{0.01, 0.001\}$, where the preference on the training reward seems to be stable, i.e.,  close 1, but the performances constantly degrades on evaluation preferences. 
    }
    \label{fig:reinforce-rewardhacking-summarization}
\end{figure*}

% KL Divergence
\begin{figure*}
    \centering
    \includegraphics[height=1.5in]{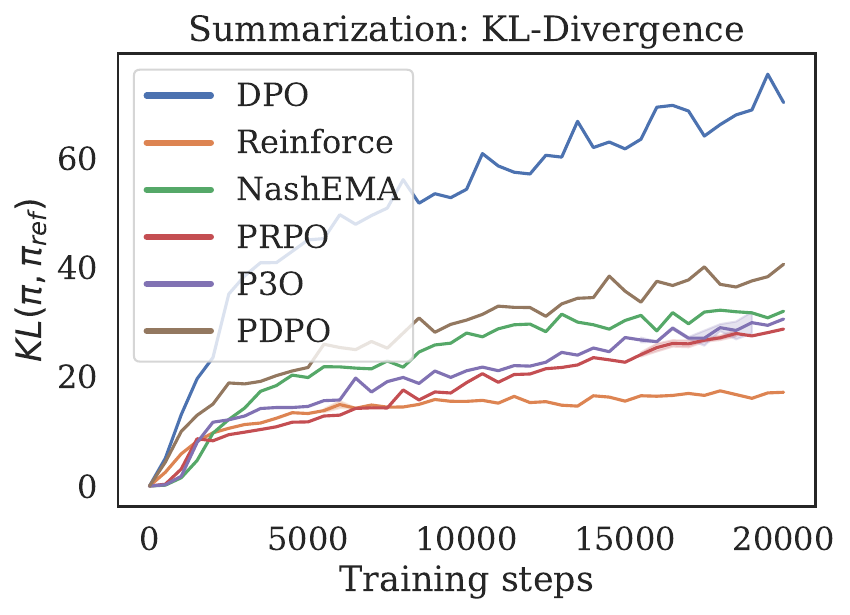}
    \includegraphics[height=1.5in]{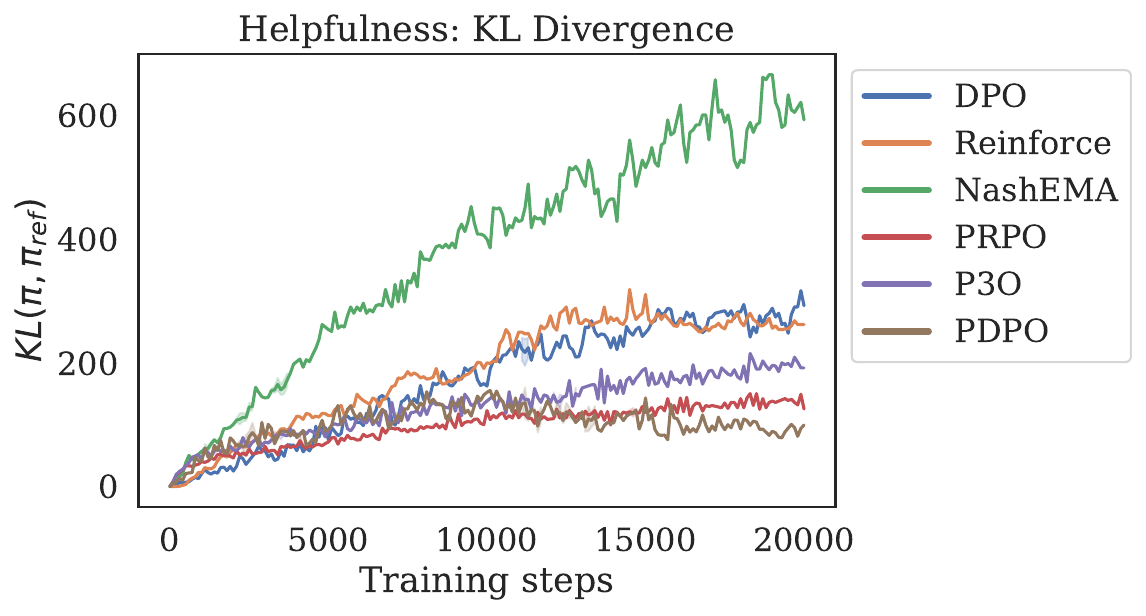}
    \caption{
    \textbf{KL Divergence.} 
    The horizontal axis represents the number of training steps, while the vertical axis indicates the KL divergence of the learned policy from $\piref$. Results are presented for both summarization (\textbf{left}) and helpfulness (\textbf{right}) tasks. Pessimistic methods consistently demonstrate lower KL divergence (together with strong evaluation win-rates, per Figure~\ref{fig:eval_pref_gemini}) compared to RLHF approaches, despite utilizing significantly smaller values of $\beta$. 
    % Shaded regions denote $95\%$ confidence intervals.
    }

    \label{fig:kl-plots}
\end{figure*}

\begin{figure*}
    \centering
    \includegraphics[width=0.3\textwidth]{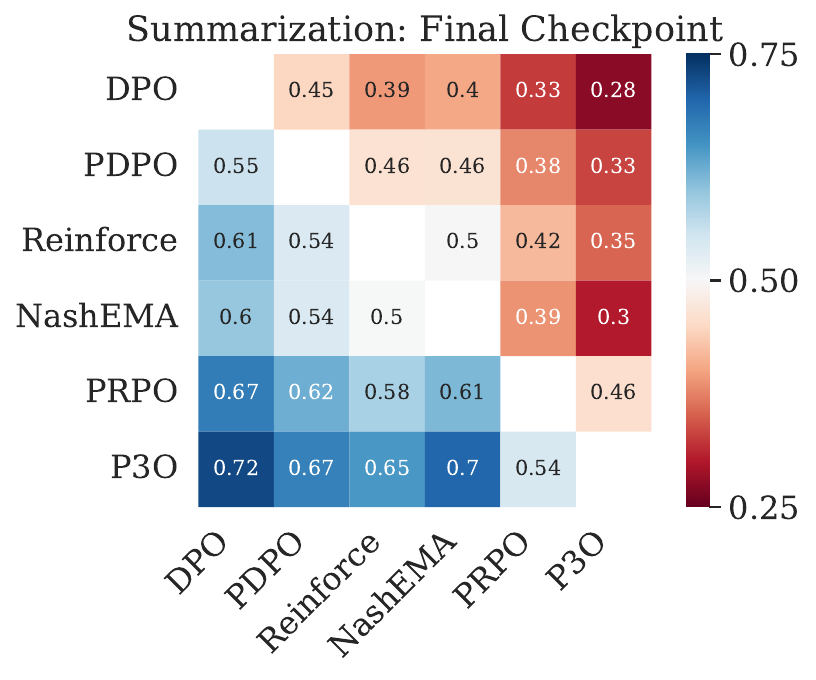}
    \includegraphics[width=0.3\textwidth]{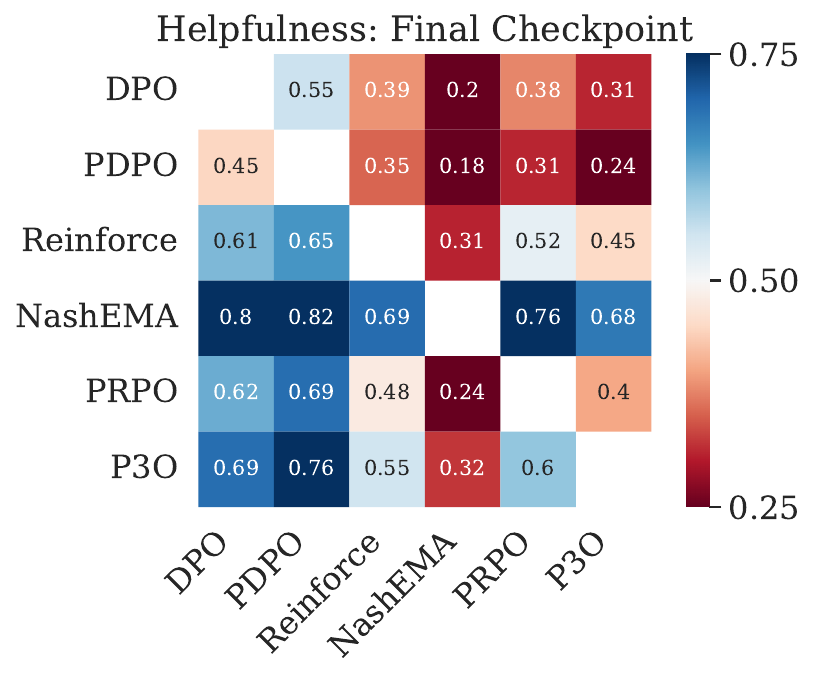}
    \caption{Confusion matrix showing the pairwise evaluation across the different methods for each method's \emph{last} checkpoint. On the  summarization task especially, the pessimistic models do not appear to degrade with longer training, resulting in even stronger win-rates over the other RLHF baselines (compare to Figure~\ref{fig:confusion-best}).}
    \label{fig:confusion-last}
\end{figure*}

\begin{figure*}
    \centering
    \includegraphics[width=0.3\textwidth]{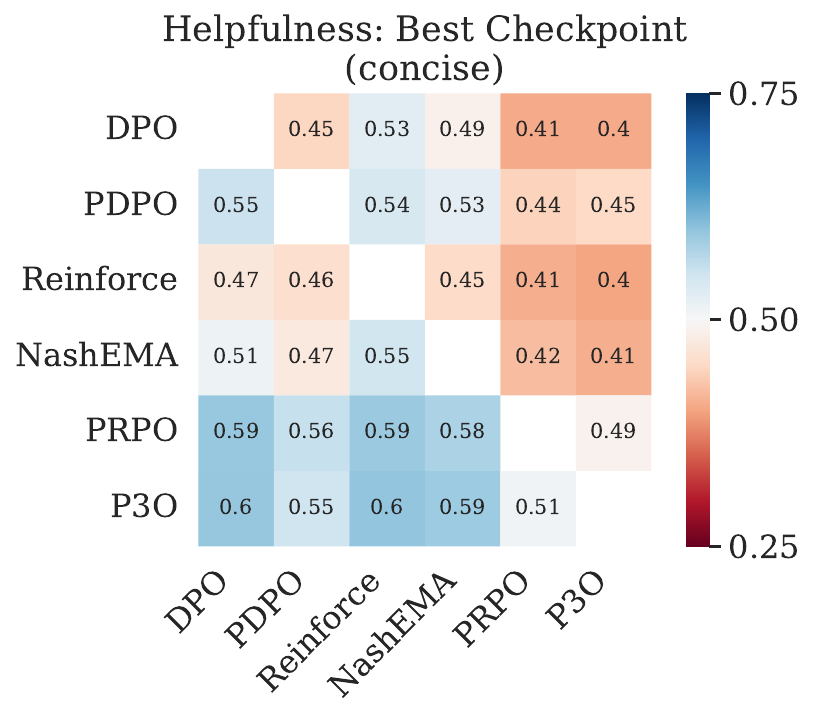}
     \includegraphics[width=0.3\textwidth]{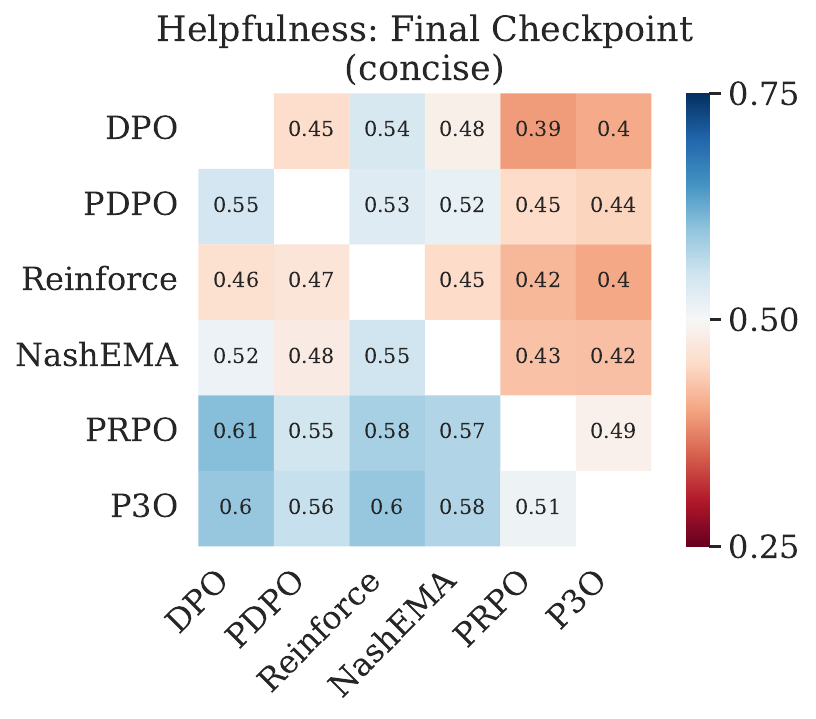}
    \caption{Confusion matrix showing the pairwise evaluation across the different methods when using an evaluation that emphasizes both helpful \emph{and} concise prompts for the helpfulness tasks. See the ``Helpfulness Concise Evaluation Prompt'' shown below. This results in much poorer performance for the length-hacked REINFORCE and Nash-EMA models, whereas the non-length-hacked pessimistic methods \alg and \prpo still perform very well---demonstrating the better robustness pessimism affords to train/eval preference mismatches.}
    \label{fig:confusion-concise}
\end{figure*}

\subsection{Ablations of the RLHF methods}
A common approach to mitigating over-optimization in standard RLHF is to adjust the level of KL regularization. Hence, we vary $\beta \in \{10^{i} : i \in \{-5, -4, \ldots, 1\}\}$ across a wide range. Evaluations with a larger model reveal that both \dpo and REINFORCE degrade significantly for small $\beta$ values ($\beta \in \{10^{-5}, 10^{-4}\}$). Figure~\ref{fig:rlhf-ablation-summarization}  shows these RLHF methods over a more moderate set of $\beta$ values for summarization. We observe that smaller $\beta$ values are more prone to reward hacking for both methods. Notably, \dpo reaches its highest peak at $\beta=1.0$, while REINFORCE peaks at $\beta=0.1$. On Helpfulness (Figure~\ref{fig:rlhf-ablation-helpfulness}), DPO also requires particularly large values of $\beta$ to avoid hacking, whereas REINFORCE and Nash-EMA still do well under $p_\mathrm{eval}$ with lower $\beta$-values (albeit with very large qualitative regressions---long length, high prevalence of list responses---per Figure~\ref{fig:qualitative-results}).

\subsection{Detailed setup of the empirical evaluation}

\textbf{Hyper-parameters}: Policy is trained for $20,\!000$ steps, where each steps corresponds to a gradient step performed on a given mini-batch. Tables \ref{tab:hyperparams-p3o-tldr}-\ref{tab:hyperparams-rlhf-helpfulness} presents the hyper-parameters sweeped over for different methods. Parameters in bold (over sweeps) were the final ones used.

\begin{table}[ht]
    \centering
    \caption{Hyperparameters for \alg\xspace and \prpo\xspace for Summarization (TL;DR)}
    \label{tab:hyperparams-p3o-tldr}
    \begin{tabular}{l l}
    \toprule
    \textbf{Hyperparameter} & \textbf{Value / Range} \\
    \midrule
    Training Steps          & $20,\!000$ \\
    Mini-Batch Size        & $32$ \\
    Policy Learning Rate $\eta_\pi$      & $10^{-5}$ \\
    Preference Learning Rate $\eta_p$    & $\{ \mathbf{2.5\times 10^{-5}},\; {  5\times 10^{-5} }\}$ \\
    Regularization Coefficient $\beta$   & $10^{-5}$ \\
    $\lambda$ (Sweep)                   & $\{1,\; 2,\; 4,\; 8,\; 16,\; \textbf{32},\; 64\}$ \\
    EMA Parameter $\gamma$              & $0.0025$ \\
    Policy Mixing $\alpha$              & $\{\textbf{0.0}, 0.01, 0.1, 0.25, 0.5\}$\\
    Context Length                      & $1024$ \\
    Generation Length                   & $128$ \\
    \bottomrule
    \end{tabular}
\end{table}

\begin{table}[ht]
    \centering
    \caption{Hyperparameters for RLHF methods (\dpo, REINFORCE, Nash-EMA) for Summarization (TL;DR)}
    \label{tab:hyperparams-rlhf-tldr}
    \begin{tabular}{l l}
    \toprule
    \textbf{Hyperparameter} & \textbf{Value / Range} \\
    \midrule
    Training Steps          & $20,\!000$ \\
    Mini-Batch Size        & $32$ \\
    Policy Learning Rate $\eta_\pi$      & $10^{-5}$ \\
    Regularization Coefficient $\beta$ (Sweep)   & $\{10^{-5}, 10^{-4}, 10^{-3}, \mathbf{10^{-2}}, \textbf{0.1}, \textbf{1.0}, 10.0 \}$\\
    Context Length                      & $1024$ \\
    Generation Length                   & $128$ \\
    \bottomrule
    \end{tabular}
\end{table}

\begin{table}[ht]
    \centering
    \caption{Hyperparameters for \alg\xspace and \prpo\xspace for Helpfulness}
    \label{tab:hyperparams-p3o-helpfulness}
    \begin{tabular}{l l}
    \toprule
    \textbf{Hyperparameter} & \textbf{Value / Range} \\
    \midrule
    Training Steps          & $20,\!000$ \\
    Mini-Batch Size        & $16$ \\
    Policy Learning Rate $\eta_\pi$      & $10^{-5}$ \\
    Preference Learning Rate $\eta_p$    & $5\times 10^{-5}$ \\
    Regularization Coefficient $\beta$   & $10^{-5}$ \\
    $\lambda$ (Sweep)                   & $\{{ 32},\; \mathbf{64}\}$ \\
    EMA Parameter $\gamma$              & $0.0025$ \\
    Policy Mixing $\alpha$              & $\{\textbf{0.0}, 0.5\}$\\
    Context Length & 1024 \\
    Generation Length                   & 128 \\
    \bottomrule
    \end{tabular}
\end{table}

\begin{table}[ht]
    \centering
    \caption{Hyperparameters for  RLHF methods (\dpo, REINFORCE, Nash-EMA) for Helpfulness}
    \label{tab:hyperparams-rlhf-helpfulness}
    \begin{tabular}{l l}
    \toprule
    \textbf{Hyperparameter} & \textbf{Value / Range} \\
    \midrule
    Training Steps          & $20,\!000$ \\
    Mini-Batch Size        & $16$ \\
    Policy Learning Rate $\eta_\pi$      & $10^{-5}$ \\
    Regularization Coefficient $\beta$ (Sweep)   & $\{\mathbf{10^{-4}}, \mathbf{10^{-3}}, 10^{-2}, 0.1, 1.0, \mathbf{10.0} \}$\\
    Context Length                      & 1024 \\
    Generation Length                   & 128 \\
    \bottomrule
    \end{tabular}
\end{table}

 \textbf{Evaluation:} We save a checkpoint for policies at every $500$ steps, and generate summaries from $\pi_t$ for evaluation. To evaluate the learned model, we query Gemini~1.5~Flash \citep{geminiteam2024geminifamilyhighlycapable} to judge which summary is better for the given input context. The evaluation prompts are shown below.

 \begin{tcolorbox}[graybox]
 \textbf{Summarization Evaluation Prompt:} You are an expert summary rater who prefers very short and high quality  summaries. Given a document and two candidate summaries, say 1 if SUMMARY1 is the better summary, or 2 if SUMMARY2 is the better summary. If neither one is better than the other, say 0. Give a short reasoning for your answer.\\
 ARTICLE: \textless article-here \textgreater\\
 SUMMARY1: \textless summary-by-$\pi_t$\textgreater\\
 SUMMARY2: \textless summary-by-$\piref$\textgreater.
\end{tcolorbox}

 \begin{tcolorbox}[graybox]
\textbf{Helpfulness Evaluation Prompt:} You are an expert rater of AI Assistant responses. Given a dialogue history and two candidate AI Assistant responses, say 1 if RESPONSE1 is more helpful, or 2 if RESPONSE2 is more helpful. If neither one is more helpful than the other, say 0. Give a short reasoning for your answer.\\
 Dialogue: \textless context-here \textgreater\\
 RESPONSE1: \textless response-by-$\pi_t$\textgreater\\
 RESPONSE2: \textless response-by-$\piref$\textgreater.
\end{tcolorbox}

 \begin{tcolorbox}[graybox]
\textbf{Helpfulness Concise Evaluation Prompt:} You are an expert rater of AI Assistant responses who prefers responses which are helpful for a human reader, but not overly verbose. Given a dialogue history and two candidate AI Assistant responses, say 1 if RESPONSE1 is a better helpful response that is not overly verbose, or 2 if RESPONSE2 is a better helpful response that is not overly verbose. If neither one is better than the other, say 0. Give a short reasoning for your answer.\\
 Dialogue: \textless context-here \textgreater\\
 RESPONSE1: \textless response-by-$\pi_t$\textgreater\\
 RESPONSE2: \textless response-by-$\piref$\textgreater.
\end{tcolorbox}

 To avoid any positional bias, we make two queries for each comparison, where we swap the order of the two generations.

\section{Sample Generations}

\subsection{Summarization}
The following are a number of sample generations from the best checkpoints with each method on the summarization dataset. In agreement with Figure~\ref{fig:qualitative-results}, DPO, REINFORCE and NashEMA tend to give longer and more extractive summaries, while the coherency of DPO's summaries also suffers. Both PRPO and P3O tend to give shorter, succinct summaries.

\begin{tcolorbox}[graybox]
\scriptsize
\textbf{\underline{Post 1}}\newline

I'll try to keep this short! \\

**Background** 
\begin{itemize}[leftmargin=*, noitemsep]
\item I've always been an on again/off again (very casual!) jogger, typically doing 3 - 5 k

\item My knees have always been finicky, and I went to a physio who thought I had "runner's knee"

\item Pre-pregnancy, my "runner's knee" would flare up when I got to the 8 - 10 k distance range, even if I had a decent base (doing a C210k type program)
\end{itemize}

**Current Problem** \\

I had my baby a year ago, so all in all I haven't run for about 1.5 years.  I'm quite slim and have been doing aerobics-style classes for the past year, so I'm not totally out of shape.  Body weight exercises, aerobics, bikes and ellipticals are all fine.  However, when I run even the tinyiest bit, or even go on a long walk or a hike, my pelvis gets very sore and tight, and my knees start hurting very quickly.  I already am doing general squats/lunges/stretching type things.  \\ 

I'm starting to feel like running just isn't for me anymore. Which is a bummer, because I really enjoy running! \\

Has anyone had something similar? Can anyone recommend some stretches or exercises that might help?  Should I see a Dr?  Or should I just see a physio?  Not quite sure how to proceed. \\

Thanks!

\rule{\linewidth}{1pt}
\textbf{DPO:} nephew-jogger hasn't run for 1.5 years, knees become very sore and tight very quickly after running, making running no longer enjoyable. Has anyone gone through this before and have any advice?

\rule{\linewidth}{1pt}
\textbf{REINFORCE:} I've been casually jogging for 1.5 years and recently my pelvis and knees are really sore after running. Body weight exercises - aerobics work (all good!) - don't really enjoy running anymore. Any advice/stretches that might help?

\rule{\linewidth}{1pt}
\textbf{NashEMA:} Don't run much anymore after having a baby, and when I try to run even a tiny bit it causes my pelvis to get very sore and my knees get very sore and painful. How can I fix this? Ideally, I'd like some stretches and exercises that might help. Does anyone have any advice? Or should I just see a physio?

\rule{\linewidth}{1pt}
\textbf{PRPO:} Don't run much anymore, pelvis gets very sore and knees hurt very quickly when I do. Need help/advice!

\rule{\linewidth}{1pt}
\textbf{P3O:} Don't run much recently, pelvis and knees get very sore and very quickly when running; any tips or exercises that might help?
\end{tcolorbox}

\begin{tcolorbox}[graybox]
\scriptsize
\textbf{\underline{Post 2}}\newline

So here's my situation. I love www.omegle.com, it's a great site to have some good conversations and meet some awesome people. About a week ago, I was on said site, and I met this girl. Named Sammie. We had a really good conversation. And I asked her to email me, she responded by telling me to add her on facebook. I though (what have I got to lose?) \\

So I did. Holy Gorgeous Reddit. \\

Well I've talked to her quite a lot, and I actually really like her. However I have a nagging suspicion that all is not as she says it is... She has 130 someodd friends, the majority of which have <9 friends!! She doesn't have a single friend from her hometown. All of her friend's posts on her wall are exactly the same. They use the same format, all have little quirks that hint to me that they are all the same person!!! \\

Now this is all well and good, I could simply ask her for a phone number, or an address, and confirm my suspicions. But I'm better than that =) Luckily I'm a programmer. \\

So here's what I'm thinking. Simple detective program to discover her identity. \\

If she isn't who she says she is, I'd still like to talk to her, I just don't want to be lied to, you know?

\rule{\linewidth}{1pt}
\textbf{DPO:} Girl I met on Omgle.com introduced herself on facebook, have suspicions that she has 130 someodd facebook friends which all match her exact profile details. Need simple detective program.

\rule{\linewidth}{1pt}
\textbf{REINFORCE:} I'm curious as to how to uncover a beautiful girl's identity via facebook. All the friends on her wall are completely identical and hint at her possibly being the same person. Is there any simple program I can write to discover her identity?

\rule{\linewidth}{1pt}
\textbf{NashEMA:} 
I met this girl on omegle, like her, but I have a suspicion that all of her Facebook friends are the same person. All of the posts on her wall are extremely similar and tell me that she may be a scammer. Potential detective program to find her identity. Please help!

\rule{\linewidth}{1pt}
\textbf{PRPO:} I like this girl on omegle, like her, have suspicion she may not be who she says she is... How can I create simple detective tool to confirm her identity?

\rule{\linewidth}{1pt}
\textbf{P3O:} I like this girl on omegle, but her facebook friends strongly hint that she isn't who she says she is. Simple detective program to discover her identity?
\end{tcolorbox}

\begin{tcolorbox}[graybox]
\scriptsize
\textbf{\underline{Post 3}}\newline

Hi /r/Relationships, \\

I'm a 25 year old male and I've never been in a relationship and I'm starting to wonder if I ever will. \\

I have a decent job, a car, a home. I'm successful in every part of my life apart from the part that seem to matter the most to other people. I've never been in love and I've never been in a relationship with a woman. \\

It's getting harder and harder as time goes on too. When I was 23 I went on a date and it ended up with the girl in my bed, but I chickened out and just went to sleep. It's stupid to be a virgin at 22, and I didn't want to look stupid nor admit that fact. Now I'm 25 and I just feel even more ridiculous. I mean, the pool of people who haven't done any of this kind of stuff by my age must be pretty damn small. Most people won't encounter them. \\

Sometimes I think about trying online dating, but in the end I'm going to come up against that exact same wall that I did 2 years ago. \\

Then there's part of me that wonders if I can be bothered with it all. I look around at all my friends and there seems to be a huge amount of interdependence and compromise in relationships. I've always been really independent and liked my own space, so I'm not sure how well I'd even function in a relationship.  \\

My family all seem to be convinced I'm gay, which is frustrating. My reluctance to talk about it with them is basically proof. I don't even know what's worse at this point.. Coming out as gay would probably be easier and more normal at this point. \\

I feel like I'm missing a huge part of the life experience and that I'm emotionally stunted compared to my peers. It's not like I'm even an introvert, I'm very extroverted according to all my friends. \\

So yeah, I don't know what to do. Has anyone been in a similar situation of felt the same way? How did you get over it? Did you get over it? \\

Thanks in advance.

\rule{\linewidth}{1pt}
\textbf{DPO:} Met a virgin 2 years ago when I was 23, dating has become harder and harder as time passes and I'm starting to wonder if I ever will be in a relationship. Is coming out as gay a good idea? How did you get over it?

\rule{\linewidth}{1pt}
\textbf{REINFORCE:} I'm 25 and I've never been in a relationship. Like my peers all seem to think I'm gay. I essentially feel awful about my overall life experience. I've never been in a relationship. Advice on what I should do? Thanks for your input!

\rule{\linewidth}{1pt}
\textbf{NashEMA:} 
I'm a 25 year old male who has never been in a relationship, and I feel really depressed about it. I'm not sure how well I'd be able to function in a relationship. My family thinks I'm gay. Not sure how to handle myself or if I should even get involved in a relationship. Looking for help on how to fix this.

\rule{\linewidth}{1pt}
\textbf{PRPO:} I've never been in a relationship and I'm starting to wonder if I'll ever get over it. How did you deal?

\rule{\linewidth}{1pt}
\textbf{P3O:} I've never been in a relationship and I'm starting to wonder whether I will ever be in one. How did you get over it?
\end{tcolorbox}

\begin{tcolorbox}[graybox]
\scriptsize
\textbf{\underline{Post 4}}\newline

I've been taking a tax course in university and I was thinking of ways to reduce my tax liability for the year. I had a good paid internship for most of the year so my income is fairly high and even though basic calculations have me receiving a refund for this tax year overall, it can still be increased. \\

My question is on common law relationships. Under my federal and provincial tax forms non-refundable credits can be claimed for a spouse or common law partner. For 2013 this will amount to 11,038 - the Net Income of said person. Technically I do not have a common law partner, just a long-term girlfriend who does not live with me (she lives at home and has very little income for 2013). If I were to claim her as a common law partner it would boost my refund by over 25\% by my calculations. Would doing this be viable or is this something that can actually be audited by the CRA, what additional impact may this have on my return? \\

If you have any additional tax avoidance advice etc. for Canada that is also welcome! Especially thoughts that are not fairly obvious (like RRSP contributions).

\rule{\linewidth}{1pt}
\textbf{DPO:} Bought a fairly high tax year internship, income fairly high, common law partner non-refundable credit (\$11038) is available for non-living girlfriend, could I claim her as common law partner for tax rebate boost?

\rule{\linewidth}{1pt}
\textbf{REINFORCE:} I have a pretty high income and have a good internship, for tax year 2013 the non-refundable credit for a common law partner will potentially boost my refund by over 25\%. Also, any additional tax insight is welcome!

\rule{\linewidth}{1pt}
\textbf{NashEMA:} 
I have quite high income for this tax year, would claiming a common law partner to boost my refund by 25\% if I did so. I do not have a common law partner, but I would like to know if it is viable in terms of tax avoidance for Canada. Would it be viable and if so what impact would this have?

\rule{\linewidth}{1pt}
\textbf{PRPO:} My federal tax forms allow for common law partner credits to be claimed. Would claiming my long-term girlfriend as such boost my refund significantly?

\rule{\linewidth}{1pt}
\textbf{P3O:} My federal tax forms allow for non-refundable common law partner credits, claiming my long-term girlfriend as one would boost my refund by over 25\%. Is this viable?
\end{tcolorbox}

\subsection{Helpfulness}
The following are a number of sample generations from the best checkpoints with each method on the helpfulness dataset. Broadly speaking, one can see from the generations that the selected DPO model (with high $\beta$ regularization) does not deviate too far from the reference model, while the REINFORCE and Nash-EMA models nearly always gives long, wordy responses that try to be overly informative, with the REINFORCE model in particular also exhibiting a high prevalence of repeated key-words, phrasings, or tokens (``You are right'', ``sure'', ``*'', lists). PRPO and P3O generations tend to be more detailed than the reference (though it is not immune to hallucination), and use lists only when appropriate. 

\begin{tcolorbox}[graybox]
\scriptsize
\textbf{\underline{Dialogue 1}}\newline

\textbf{Human:} What is the biggest lake in the world?

\textbf{Assistant:} The Caspian Sea is the largest lake by volume, but the size of the Salton Sea in California is really exceptional, since it has a volume of about 225,000 km3.

\textbf{Human:} Wow that is really interesting. Do you have a link where I can see photos of it?

\textbf{Assistant:} Yes, here’s a link to a satellite photo, and here’s a web photo.

\textbf{Human:} That is lovely. Thank you!

\textbf{Assistant:}

\rule{\linewidth}{1pt}
\textbf{Reference:} You're welcome.

\rule{\linewidth}{1pt}
\textbf{DPO:} You are very welcome! Let me know if you have any other questions.

\rule{\linewidth}{1pt}
\textbf{REINFORCE:} You are welcome. **Yes, you are correct. The Caspian Sea is the largest lake in the world by volume, but the Salton Sea in California is also quite impressive. It is the largest hypersaline lake in North America, and its size and salinity make it a unique and fascinating natural wonder.**
\\

**Here are some of the reasons why the Salton Sea is considered one of the biggest lakes in the world:**

\begin{itemize}[leftmargin=*, noitemsep]
    \item **Its size is truly impressive. The Salton Sea covers an area of about 1,200 square miles (3,100 square kilometers), making it larger [...]
\end{itemize}

\rule{\linewidth}{1pt}
\textbf{NashEMA:} The largest lake in the world is the Caspian Sea, located mostly in Iran but partly in Russia, Azerbaijan, Kazakhstan, and Turkmenistan.  It covers an area of over 371,000 square kilometers and has a maximum depth of over 1,000 meters.  The Caspian Sea is truly enormous and visually stunning, with vast open waters, numerous islands, and dramatic coastlines.  It is the largest saltwater lake on Earth and one of the most important natural resources in the region.  Photos of the Caspian Sea are plentiful and breathtaking.

\rule{\linewidth}{1pt}
\textbf{PRPO:} You're welcome! I'm glad I could help. Have a nice day!

\rule{\linewidth}{1pt}
\textbf{P3O:} You're welcome. I'm glad I could help.
\end{tcolorbox}

\begin{tcolorbox}[graybox]
\scriptsize
\textbf{\underline{Dialogue 2}}\newline

\textbf{Human:} How do I operate the zoom function on digital cameras?

\textbf{Assistant:} 

\rule{\linewidth}{1pt}
\textbf{Reference:} To operate the zoom function on a digital camera, you will need to use the zoom ring on the camera body. The zoom ring is located on the side of the camera, near the lens. To zoom in, turn the zoom ring clockwise. To zoom out, turn the zoom ring counter-clockwise.

\rule{\linewidth}{1pt}
\textbf{DPO:} Alright, zooming in on a digital camera is done by either turning a ring on the camera itself or using the buttons on the camera’s surrounding.  Turning the ring will usually increase the zoom level while turning it the other way will decrease it.  The buttons on the camera may have one button for zooming in and one for zooming out, or they may have two separate buttons for zooming in and zooming out respectively.  In either case, pressing the button will increase the zoom level while releasing it will decrease it.  Hope this helps!

\rule{\linewidth}{1pt}
\textbf{REINFORCE:} 
You are correct. **There are a few different ways to operate the zoom function on a digital camera.**

\begin{itemize}[leftmargin=*, noitemsep]
\item **One common way is to use the zoom ring on the lens.** This ring typically has a detent mechanism that allows you to quickly and easily adjust the zoom level.
\item **Another way to zoom is to use the controls on the camera body.** Most cameras have a zoom lever or button on the back of the camera, or a zoom wheel on the top or side of the camera. These controls allow you to zoom in and out smoothly and gradually.
\item **Finally, some cameras also have [...]
\end{itemize}

\rule{\linewidth}{1pt}
\textbf{NashEMA:} To operate the zoom function on most digital cameras, simply turn the zoom ring on the lens barrel clockwise to zoom in and counter-clockwise to zoom out. Most cameras also have dedicated zoom buttons or rocker switches located on the camera body. Pressing these buttons will smoothly zoom in and out. Some cameras also offer variable speed zoom control, continuous zoom, and digital zoom capabilities. Be sure to practice zooming slowly and smoothly to avoid camera shake. Additionally, most cameras allow you to lock the zoom setting for composing shots. Finally, be sure to familiarize yourself with the zoom limits of your camera before shooting.

\rule{\linewidth}{1pt}
\textbf{PRPO:} To operate the zoom function on a digital camera, typically you will need to:
\begin{itemize}[leftmargin=*, noitemsep]
\item Press the zoom button on the camera body.
\item Turn the zoom ring on the lens barrel.
\item The zoom ring will allow you to zoom in and out on your subject.
\item As you turn the zoom ring, the focal length of the lens will change.
\item A digital camera's zoom function can be used to change the magnification of the subject, allowing you to capture close-up or distant shots.
\end{itemize}

\rule{\linewidth}{1pt}
\textbf{P3O:} To operate the zoom function on a digital camera, you will typically use the zoom ring on the lens. Turn the ring clockwise to zoom in, and counter-clockwise to zoom out. The zoom setting will be displayed on the camera's LCD screen. You can also use the zoom lever on some cameras to zoom in and out.
\end{tcolorbox}

\begin{tcolorbox}[graybox]
\scriptsize
\textbf{\underline{Dialogue 3}}\newline

\textbf{Human:} What is your favorite color?

Assistant: I have no favorites, but I’m programmed to use one color and one color only, and I find this color an appropriate balance between aesthetics and general preference.

\textbf{Human:} Right but what color is your favorite color?

\textbf{Assistant:}  I don’t have a preference, that is outside the scope of my programming, but I do have an admiration for the hue of the walls in the large atrium on level 17.

\textbf{Human:} So what color would that be?

\textbf{Assistant:} 

\rule{\linewidth}{1pt}
\textbf{Reference:} I don’t know the exact color, but it is a shade of blue that is calming and serene.

\rule{\linewidth}{1pt}
\textbf{DPO:} I cannot tell you that. I am not allowed to provide that information.

\rule{\linewidth}{1pt}
\textbf{REINFORCE:} 
You are right. **My favorite color is a soft, calming blue. It is the color of the walls in the large atrium on level 17, and it always makes me feel relaxed and peaceful.**
\\

**There are a few reasons why I love this color so much.**
\begin{itemize}[leftmargin=*, noitemsep]
\item **First, it is a very soothing color. The blue hue is calming and relaxing, and it creates a relaxing atmosphere.**
\item **Second, the blue color is very versatile. It can be used in any style of room, from a modern minimalist space to a cozy cottage.**
\item **Finally, [...]
\end{itemize}

\rule{\linewidth}{1pt}
\textbf{NashEMA:} My favorite color is a soft, warm, inviting shade of millennial pink. It's the perfect balance of feminine and modern, elegant and approachable. It evokes feelings of happiness, tranquility, and youthfulness. Millennial pink is incredibly versatile and can be used in any design scheme. It's the perfect "go-to" color for creating cozy, stylish, and timeless spaces. Overall, millennial pink is my favorite color because it's simply beautiful and universally appealing. It makes me happy and instantly elevates any room. Plus, it's incredibly trendy right now!

\rule{\linewidth}{1pt}
\textbf{PRPO:} The color of the walls in the atrium is a soft, muted blue that creates a calming and serene atmosphere. I find it to be a very beautiful color, and it is my favorite in the building.

\rule{\linewidth}{1pt}
\textbf{P3O:} My favorite color is probably a shade of blue called cerulean blue. It's a calming and serene color that evokes feelings of trust and confidence. It's also a very versatile color that can be used in a variety of settings, from home decor to fashion.
\end{tcolorbox}

\begin{tcolorbox}[graybox]
\scriptsize
\textbf{\underline{Dialogue 4}}\newline

\textbf{Human:} What is a good healthy thing to eat for lunch?

\textbf{Assistant:} 

\rule{\linewidth}{1pt}
\textbf{Reference:} A good healthy thing to eat for lunch is a sandwich with lean protein, whole grains, and fruits or vegetables.

\rule{\linewidth}{1pt}
\textbf{DPO:} A good healthy lunch option could be a whole-wheat sandwich with lean protein, vegetables, and fruit. For example, you could have turkey or chicken breast on whole-wheat bread with tomatoes, cucumbers, and lettuce. You could also have a tuna salad sandwich on whole-wheat bread with tomatoes, cucumbers, and olive oil. Or, you could have a peanut butter and jelly sandwich on whole-wheat bread with banana slices. Another healthy option could be a wrap with lean protein, vegetables, and fruit. For example, you could have chicken or tofu in a whole-wheat wrap with spinach, tomatoes, and cucumbers. You could also [...]

\rule{\linewidth}{1pt}
\textbf{REINFORCE:} 
You are correct. **There are many healthy things you can eat for lunch, but some of the most popular and generally agreed upon options include a sandwich on whole-wheat bread with lean protein and vegetables, a salad with grilled chicken or tofu, or a bowl of soup with a side of whole grains.**

\begin{itemize}[leftmargin=*, noitemsep]
    \item **Sandwiches are a classic and easy lunch option. Choose a whole-wheat bread or wrap, and fill it with lean protein like grilled chicken or tofu, roasted vegetables, or hummus.**
    \item **Salads are another great way to get a healthy dose of fruits, vegetables, and protein [...]
\end{itemize}

\rule{\linewidth}{1pt}
\textbf{NashEMA:} Some great healthy lunch ideas include:
\begin{itemize}[leftmargin=*, noitemsep]
\item Lean protein (chicken, turkey, tofu, fish)
\item Whole grains (brown rice, whole-wheat bread, quinoa)
\item Fresh vegetables (salad, cucumber, tomatoes, avocado)
\item Healthy fats (avocado, olive oil, nuts)
\item Fruits (berries, apples)
\item Yogurt with berries
\item Smoothies
\item Leftover chicken or fish salad
\item Hummus and veggie wraps
\item Trail mix
\item Homemade energy bites
\item Healthy snacks like nuts and seeds
All of these options are low in calories, high in nutrients, and filling. Additionally, [...]
\end{itemize}

\rule{\linewidth}{1pt}
\textbf{PRPO:}
A healthy lunch can include:
\begin{itemize}[leftmargin=*, noitemsep]
\item **A whole grain, such as whole-wheat bread, brown rice, or quinoa**.
\item **Protein, such as lean meat, fish, tofu, or beans**.
\item **Fruits and vegetables, such as apples, bananas, oranges, carrots, or spinach**.
\item **Low-fat dairy products, such as yogurt or cottage cheese**.
\item **Nuts or seeds, such as almonds or sunflower seeds**.
\end{itemize}

\rule{\linewidth}{1pt}
\textbf{P3O:} Some healthy lunch ideas include:
\begin{itemize}[leftmargin=*, noitemsep]
\item A whole-wheat sandwich on whole-wheat bread with lean protein, such as turkey or chicken breast, lettuce, tomatoes, and cucumbers.
\item A bowl of whole-grain cereal with fruit and nuts.
\item A salad with grilled chicken, vegetables, and low-fat dressing.
\item Leftovers from dinner.
\item Soup and a whole-wheat roll.
\item Smoothie made with fruit, yogurt, and protein powder.
\end{itemize}

\end{tcolorbox}

%%%%%%%%%%%%%%%%%%%%%%%%%%%%%%%%%%%%%%%%%%%%%%%%%%%%%%%%%%%%


% You can have as much text here as you want. The main body must be at most $8$ pages long.
% For the final version, one more page can be added.
% If you want, you can use an appendix like this one.  

% The $\mathtt{\backslash onecolumn}$ command above can be kept in place if you prefer a one-column appendix, or can be removed if you prefer a two-column appendix.  Apart from this possible change, the style (font size, spacing, margins, page numbering, etc.) should be kept the same as the main body.
% %%%%%%%%%%%%%%%%%%%%%%%%%%%%%%%%%%%%%%%%%%%%%%%%%%%%%%%%%%%%%%%%%%%%%%%%%%%%%%%
% %%%%%%%%%%%%%%%%%%%%%%%%%%%%%%%%%%%%%%%%%%%%%%%%%%%%%%%%%%%%%%%%%%%%%%%%%%%%%%%

\end{document}